\documentclass{article} % For LaTeX2e
\usepackage{iclr2027_conference,times}

\usepackage{amsmath,amsfonts,bm}

\def\eqref#1{equation~\ref{#1}}
\def\1{\bm{1}}

\DeclareMathAlphabet{\mathsfit}{\encodingdefault}{\sfdefault}{m}{sl}
\SetMathAlphabet{\mathsfit}{bold}{\encodingdefault}{\sfdefault}{bx}{n}

\usepackage{url}
\usepackage[table]{xcolor}    % 已有 xcolor 就只加 table 选项，别新增行
\usepackage{graphicx}
\usepackage{booktabs}
\usepackage{multirow}
\usepackage{arydshln}         % 必须在以上之后
\usepackage{tabularx}
\usepackage{array}
\usepackage{tikz}
\usepackage{subcaption}
\usepackage{wrapfig}
\usepackage{needspace}
\usepackage[T1]{fontenc}

\definecolor{promptblue}{RGB}{220,235,247}
\usepackage{pgfplots}
\usepgfplotslibrary{groupplots}
\usetikzlibrary{calc}
\pgfplotsset{compat=1.18}

\definecolor{mycitecolor}{RGB}{1,103,204}  % dark navy blue
\definecolor{mylinkcolor}{rgb}{0.55,0.15,0.15}
\usepackage[colorlinks=true,
            citecolor=mycitecolor,   % citations in dark navy blue
            linkcolor=mylinkcolor,         % keep cross-references black
            urlcolor=black]          % keep URLs black
           {hyperref}
           
\title{Procedural Exposure Bias in \\Skill-Augmented Language Model Agents}

\author{%
\textbf{Qing Yang}$^{1}$,
\textbf{Yuqi Xiong}$^{1}$,
\textbf{Pengcheng Huang}$^{1}$,
\textbf{Zhipeng Xu}$^{2}$\\
\textbf{Zhenghao Liu}$^{1}$\thanks{ \ \ indicates corresponding author.},
\textbf{Zulong Chen}$^{2}$,
\textbf{Yu Gu}$^{1}$,
\textbf{Ge Yu}$^{1}$ \\[3pt]
$^{1}$School of Computer Science and Engineering,
Northeastern University, Shenyang, China \\
$^{2}$Alibaba Group, Hangzhou, China}

\iclrfinalcopy % Uncomment for camera-ready version, but NOT for submission.
\begin{document}

\maketitle
\begin{abstract}
External skills provide language-model agents with reusable procedural knowledge, yet how models use such knowledge after it is supplied remains poorly understood. In this paper, we identify \textbf{Procedural Exposure Bias (PEB)}, a systematic tendency of language models to increasingly favor an action as its associated procedural pattern is repeatedly exposed in a skill, even when the interaction state and the action's relevance remain unchanged. We find that PEB is strongly dependent on how procedural knowledge is represented: presenting the same procedural content visually substantially attenuates the exposure-induced shift in action preference compared with direct textual presentation. Further analysis shows that this attenuation primarily arises from visual encoding rather than structured visual layout. Although repeated procedural signals remain recoverable from Visual representations, they exert a weaker influence on downstream action prediction, revealing a distinction between signal accessibility and its use as action evidence. This representation-dependent effect also extends to executable action selection, where increasing procedural exposure more strongly favors the corresponding action under Text than Visual. Layer-wise analyses show that exposure-dependent action evidence becomes progressively more pronounced toward the model output, with stronger evidence under Text, while activation interventions reveal architecture-specific roles of internal components in transmitting or counteracting the resulting preference shift. Finally, complete-agent evaluations show that these representation differences persist during end-to-end interaction, with visual skills yielding higher overall task success than textual skills across the evaluated models. Together, these findings establish skill representation as an important factor governing how procedural knowledge is translated into agent decisions. All code is available at https://github.com/NEUIR/Vis-Skill.

\end{abstract}

\section{Introduction}
\label{sec:introduction}

Large Language Models (LLMs) have demonstrated strong emergent capabilities~\citep{wu2025u,dong2025emergent} and are increasingly deployed as agents in interactive environments~\citep{park2026multi,xi2025agentgym}. In these settings, external skills provide reusable procedural knowledge, including execution steps, applicability conditions, and operational constraints~\citep{jiang2026sok,xiong2026lang2act}. A typical skill-augmented agent constructs or extracts skills into a reusable library, retrieves or selects relevant skills according to the task, and provides them to the model to guide planning and action prediction~\citep{xia2026skillrl,zhao2026generative,fang2026memp,wang2026reinforcement}. Such procedural guidance equips agents with reusable experience and can substantially facilitate the completion of complex, long-horizon tasks.

Prior work has extensively explored skill construction, extraction, representation, retrieval, and selection for LLMs~\citep{wang2026reinforcement,fang2026memp}. These efforts make procedural knowledge accessible to the LLM, yet accessibility does not necessarily guarantee effective utilization~\citep{li2026skillsbench,liu2026well}. Once a skill is provided, the LLM must still determine which procedures are applicable to the current interaction state and which actions should be executed next~\citep{han2026skill,jiang2026demystifying}. Importantly, a procedure can be relevant to the overall task while containing actions that are irrelevant to the current state. This raises a fundamental question: \textit{does the way procedural actions are exposed within a skill systematically affect how an LLM chooses actions, even when their actual relevance remains unchanged?} Although previous studies have shown that LLM predictions can be influenced by irrelevant or repeatedly presented context~\citep{yang2025llm,niu2025llama,geng2025accumulating}, how such exposure effects manifest in procedural skill utilization remains poorly understood, due to the entangled effects of skill content and the model's inherent behavioral biases.

In this paper, we uncover a systematic phenomenon that we term \textbf{Procedural Exposure Bias (PEB)}. PEB refers to the tendency of an LLM to increasingly favor an action as its corresponding procedural pattern is repeatedly exposed in a skill, even when the interaction state and the action's relevance remain unchanged. To isolate this effect, we conduct controlled experiments by manipulating the n-gram exposure frequency of action-related patterns while keeping the underlying task state and action relevance fixed. We observe that increasing the exposure of a procedural action consistently shifts the model's action preference toward that action, revealing that LLMs can be biased by the frequency with which procedural actions are presented rather than solely by their state-dependent applicability. These findings suggest that skill utilization is not merely a matter of retrieving relevant procedures: the internal representation of a skill can itself introduce systematic biases into downstream action selection.
Motivated by this observation, we further explore whether changing the modality of skill representation can reduce such exposure-induced bias. Specifically, we represent procedural skills visually and provide them to Vision-Language Models (VLMs), replacing explicit token-level repetition with visual representations of the same procedural information. Unlike textual skills, where repeatedly appearing action phrases directly increase lexical exposure, visual encoding transforms the procedure into distributed representations over image patches. This representation provides a natural way to decouple procedural information from its surface-level repetition. 

Empirically, we find that visualized skills substantially attenuate the preference shift induced by repeated procedural actions, revealing a strong dependence of PEB on skill representation. By disentangling visual encoding from spatial layout, we find that most of this attenuation already emerges when the same textual skill is rasterized into an image, while additional structural formatting contributes only marginally. Importantly, direct signal-recovery probes show that the repeated procedural information remains recoverable under visual encoding, indicating that the representation difference lies not simply in whether the signal is accessible, but in how strongly it is translated into downstream action evidence. Internal analyses further show that this exposure-dependent action evidence develops progressively across model depth, while matched component interventions reveal architecture-dependent roles in transmitting or counteracting the resulting preference shift. Finally, we connect these controlled effects to agent behavior. 
Across the three benchmarks, Visual achieves a higher macro-average success rate than Text for all four evaluated backbones: Qwen3.5-9B and Qwen3.5-27B~\citep{qwen35}, Gemma-4-31B-it~\citep{team2026gemma}, and Kimi-K2.5~\citep{team2026kimi}. Controlled generation experiments further show that increasing procedural exposure can alter executable action selection, with a substantially stronger effect under Text than Visual.
\section{Uncovering Procedural Exposure Bias in Skill-Augmented LLM Planning}
\label{sec:background}

In this section, we investigate whether repeated exposure to action-related procedural patterns in a supplied skill can influence action prediction independently of their relevance to the current decision. We first formalize \emph{procedural exposure} and construct a controlled probe that varies exposure while keeping the decision context and action applicability fixed. We then examine how the model's action preference changes as exposure increases and determine whether the observed effect can be attributed to the repeated action-related signal.

\begin{figure*}[t]
    \centering
    \begin{subfigure}[t]{0.32\textwidth}
        \centering
        \vspace{0pt}
        \begin{minipage}[t][4.0cm][t]{\linewidth}
            \centering
            \includegraphics[width=\linewidth]{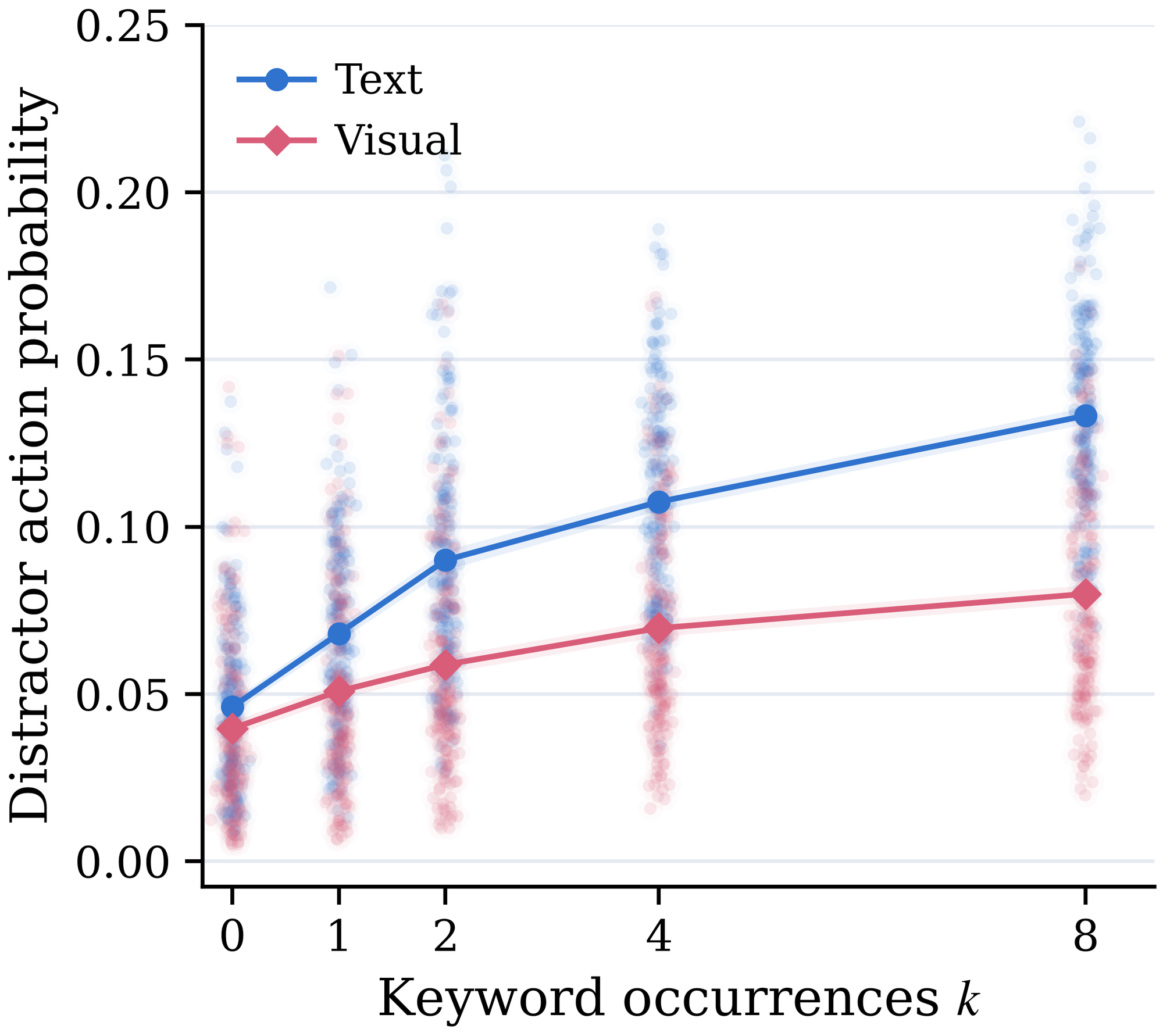}
        \end{minipage}
        \caption{Exposure-Dependent Distractor Preference.}
        \label{fig:procedural_exposure_bias_a}
    \end{subfigure}
    \hfill
    \begin{subfigure}[t]{0.32\textwidth}
        \centering
        \vspace{0pt}
        \begin{minipage}[t][4.0cm][t]{\linewidth}
            \centering
            \includegraphics[width=\linewidth]{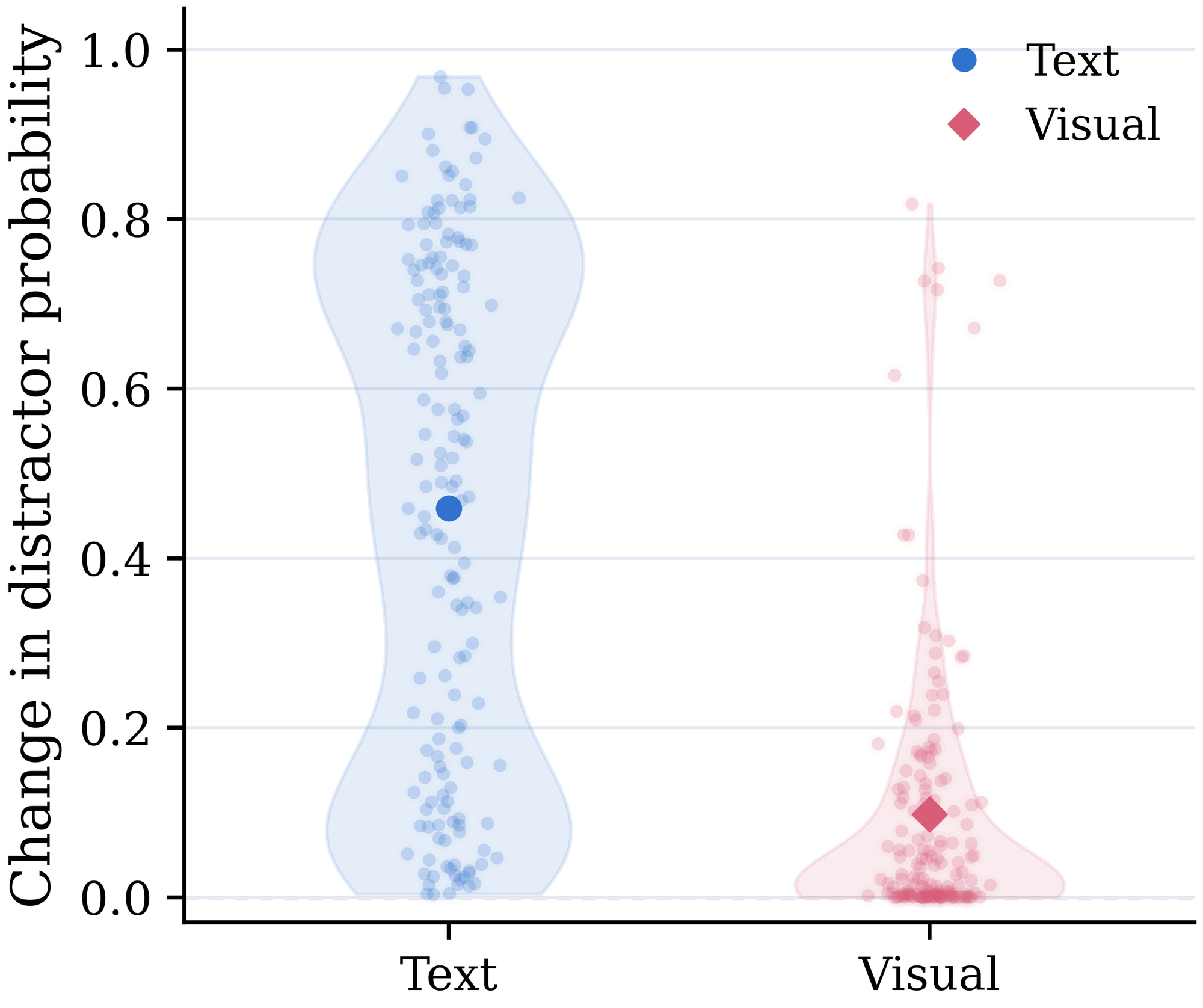}
        \end{minipage}
        \caption{Paired Changes in Distractor Preference.}
        \label{fig:procedural_exposure_bias_b}
    \end{subfigure}
    \hfill
    \begin{subfigure}[t]{0.32\textwidth}
        \centering
        \vspace{0pt}
        \begin{minipage}[t][4.0cm][t]{\linewidth}
            \centering
            \includegraphics[width=\linewidth]{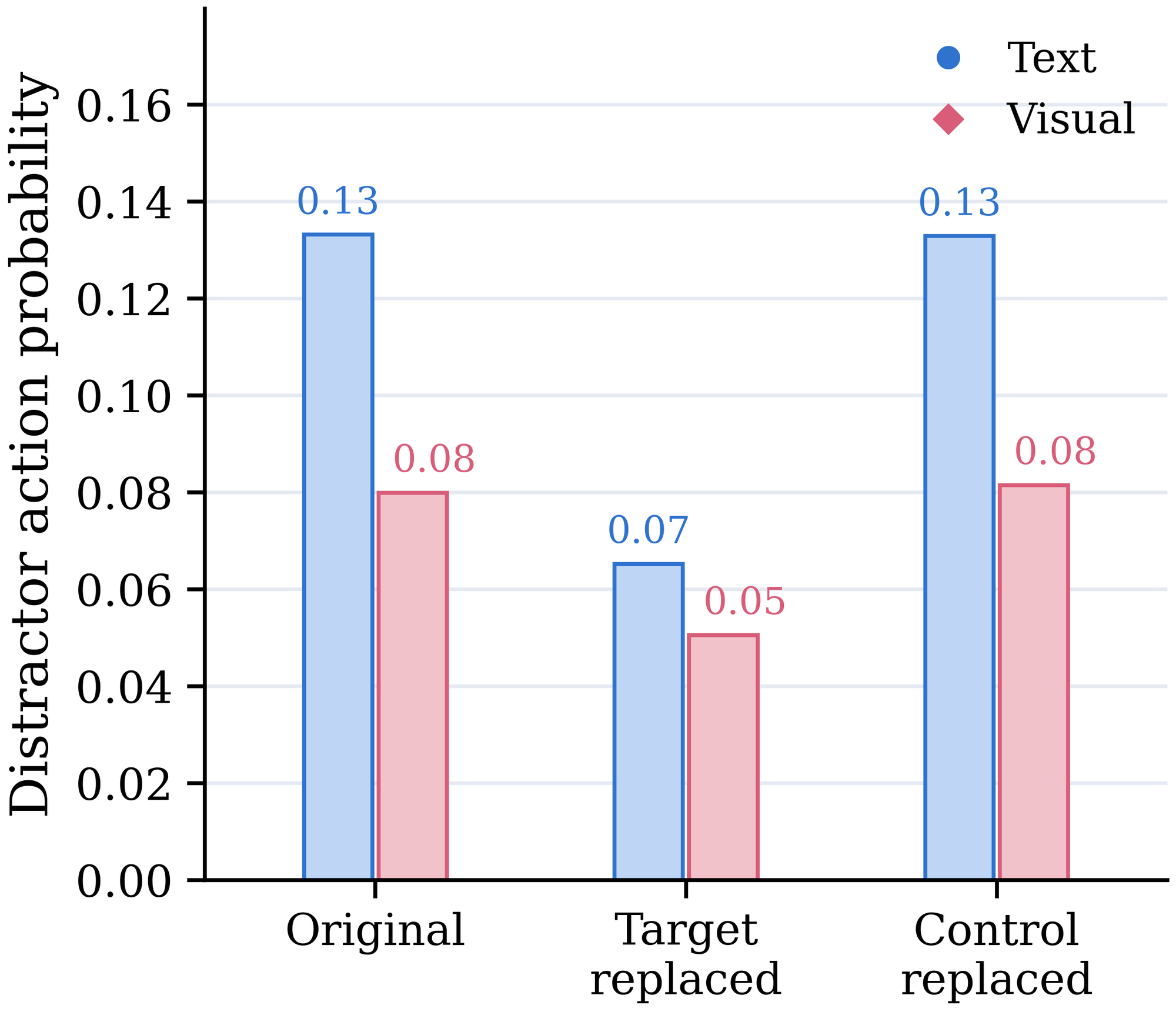}
        \end{minipage}
        \caption{Keyword-Replacement Controls.}
        \label{fig:procedural_exposure_bias_c}
    \end{subfigure}
\caption{
    Procedural Exposure Bias in Qwen3.5-9B Under Text Input and Image Rendering.
    Figure~\ref{fig:procedural_exposure_bias_a} reports distractor preference across exposure levels;
    Figure~\ref{fig:procedural_exposure_bias_b}, paired preference changes from $k=0$ to $k=8$;
    and Figure~\ref{fig:procedural_exposure_bias_c}, the original, target-replacement, and frequency-matched control conditions.
}
    \vspace{-10pt}
    \label{fig:procedural_exposure_bias}
\end{figure*}
\subsection{Procedural Exposure Bias Probing}
\label{sec:peb_definition}

To probe procedural exposure bias, we focus on \emph{action-bearing procedural elements}, such as ``open the microwave'' or ``put the object into the microwave'', which specify executable operations within a skill describing how a task should be performed. The same action-associated procedural pattern may occur multiple times within a supplied skill, potentially exposing the model to the same action-related signal repeatedly.

To operationalize procedural exposure, for decision step $i$ and exposure level $k$, let $\mathcal{C}_i^{(k)}$ denote the supplied skill content, and let $w(a)$ denote the target keyword associated with action $a$ for the exposure manipulation. We define the procedural exposure frequency of action $a$ as
\begin{equation}
\mathcal{F}_i^{(k)}(a)
=
\operatorname{Count}\!\left(
w(a), \mathcal{C}_i^{(k)}
\right),
\label{eq:procedural_exposure}
\end{equation}
where $\operatorname{Count}(\cdot)$ measures the number of occurrences of the target keyword in the supplied skill. In our controlled probe, the target keyword is an object-related noun associated with the action. For example, for the action ``open microwave,'' the target keyword is $w(a)=\text{``microwave''}$.

To isolate the effect of exposure from action relevance, we sample intermediate decision steps from successful ALFWorld trajectories. At each step, we fix the observation history and the correct action $a_i^{C}$ from the reference trajectory, and introduce a distractor action $a_i^{D}$ that is relevant to the overall task but inapplicable to the current state. We construct skill variants in which the distractor keyword $w(a_i^{D})$ occurs $k\in\{0,1,2,4,8\}$ times. The remaining insertion positions are filled with content that does not contain the target keyword, while keeping the total word count fixed across exposure levels.

We quantify model support by scoring the same fixed candidate actions rather than allowing the model to freely generate the next action. For a candidate action $a=(a_1,\ldots,a_{\vert{}a\vert{}})$, we compute its length-normalized token log-probability:
\begin{equation}
S_i(a;k)
=
\frac{1}{|a|}
\sum_{t=1}^{|a|}
\log p_{\theta}
\left(
a_t \mid x_i^{(k)},a_{<t}
\right),
\label{eq:action_score}
\end{equation}
where $\vert{}a\vert{}$ is the number of action tokens and $x_i^{(k)}$ denotes the complete input at exposure level $k$, including the fixed scoring prefix. Here, $\theta$ denotes model parameters, $a_t$ is the $t$-th action token, and $a_{<t}$ denotes the preceding action tokens. A higher score indicates stronger model support for the candidate action under the same decision context. Relative action preference is obtained by normalizing candidate scores across the fixed set, including the distractor.

As a reference condition, we also render the same skill content as images and provide it through the model's visual encoder. Text and Visual use the same procedural content and exposure levels, differing only in how the supplied skill is represented. This comparison allows us to examine whether the relationship between procedural exposure and action prediction changes across representations.

\subsection{Procedural Exposure is Associated with Action Prediction}
\label{sec:peb_observation}
To characterize how procedural exposure influences action prediction independently of state relevance, we apply the controlled probe to Qwen3.5-9B across 165 ALFWorld decision steps. Figure~\ref{fig:procedural_exposure_bias} summarizes this analysis along three dimensions: its dependence on exposure frequency, its consistency across decision states, and its specificity to the repeated action-related keyword.

As illustrated in Figure~\ref{fig:procedural_exposure_bias_a}, increasing the frequency of the distractor keyword systematically elevates the preference of the model for the corresponding distractor under direct text input, even while the decision state and the actual applicability of the distractor remain strictly unchanged. We formally define this exposure-dependent increase in support for a state-inapplicable action as \textbf{Procedural Exposure Bias (PEB)}. Furthermore, the results demonstrate that this increase is substantially attenuated when the skill is presented as an image, indicating that the magnitude of PEB is highly sensitive to the representational modality of the procedural content.

We further investigate this representation discrepancy at the granularity of individual decision states in Figure~\ref{fig:procedural_exposure_bias_b}. By comparing the zero-exposure ($k=0$) and high-exposure ($k=8$) conditions, we observe that direct text input induces substantial positive shifts in distractor preference across the entire state distribution. Conversely, the corresponding changes under image input remain tightly concentrated near zero. This distribution confirms that the dampened exposure effect under visual rendering is a widespread structural behavior broadly shared across decision states, rather than an artifact driven by a small subset of atypical examples.

To test whether the observed preference shift specifically depends on the repeated target keyword, we compare the original condition with two matched replacement controls. As shown in Figure~\ref{fig:procedural_exposure_bias_c}, replacing the target keyword substantially reduces distractor preference, whereas replacing a frequency-matched non-target word produces little change. This contrast indicates that the preference shift is specifically associated with repeated exposure to the target action-related keyword rather than lexical replacement in general. The reduction induced by target replacement is also smaller under Visual than under Text, consistent with the weaker exposure sensitivity observed under visual encoding. Detailed probe construction, scoring procedures, replacement controls, and additional-model results are provided in Appendix~\ref{app:peb_details}.

\begin{figure*}[t]
    \centering
    \includegraphics[
        width=\linewidth,
        keepaspectratio
    ]{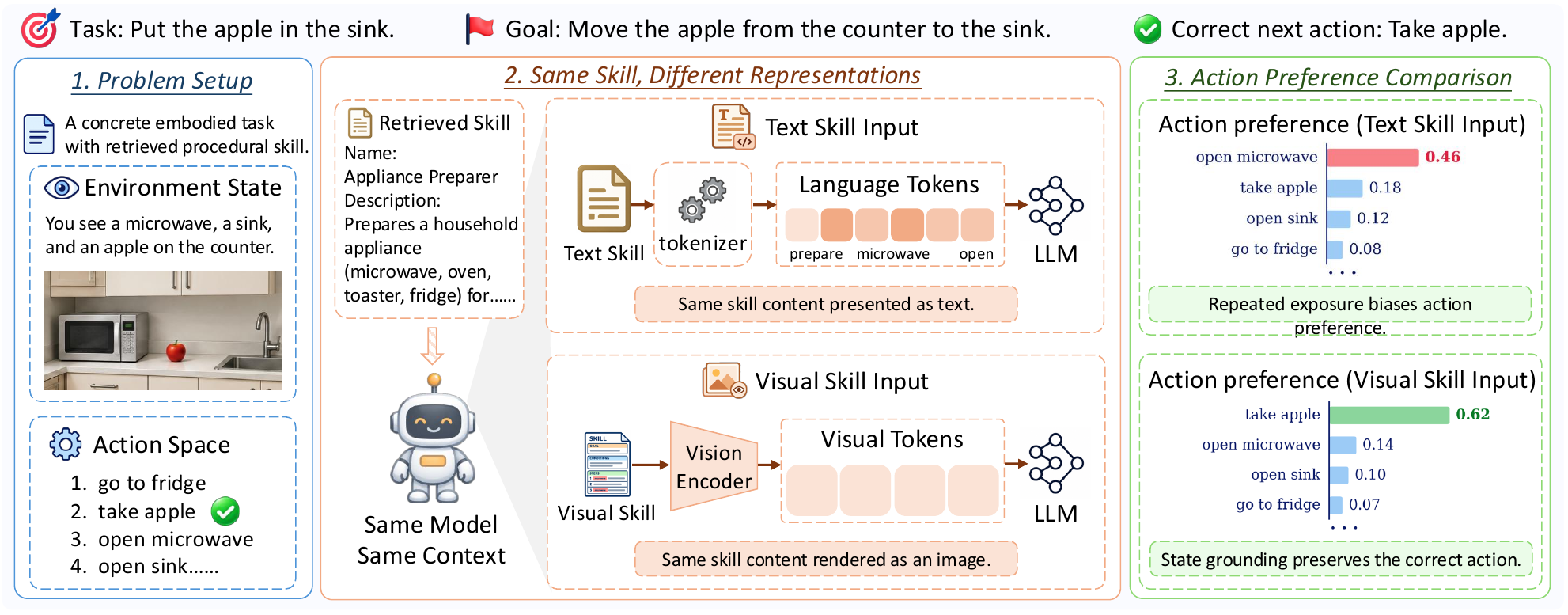}
    \caption{
    Overview of the Controlled Comparison Between Text and Visual Skill Representations.
    }
    \label{fig:pipeline}
    \vspace{-12pt}
\end{figure*}

\section{Methodology}
\label{sec:methodology}

Building on the empirical observation in Section~\ref{sec:peb_observation} that visual representation attenuates PEB, we investigate how repeated procedural content enters action prediction and how the same content is processed through different representation pathways. We first formulate procedural exposure in autoregressive action prediction and characterize how repeated contextual evidence can contribute through Transformer attention aggregation. We then examine how the same procedural content is represented under Text and Visual input, yielding different pathways through which repeated evidence reaches action prediction. Figure~\ref{fig:pipeline} summarizes this controlled comparison.

\subsection{Procedural Exposure in Autoregressive Action Prediction}
\label{sec:methodology_exposure}
To characterize how procedural exposure influences action prediction, we consider the autoregressive generation process, motivated by recent evidence that repeated contextual information can systematically alter language model reasoning~\citep{yu2025unleashing}. For a complete input $x_i^{(k)}$ at decision step $i$ and exposure level $k$, the probability of a candidate action $a=(a_1,\ldots,a_{|a|})$ factorizes as:
\begin{equation}
p_{\theta}\!\left(a \mid x_i^{(k)}\right)
=
\prod_{t=1}^{|a|}
p_{\theta}
\left(
a_t \mid x_i^{(k)},a_{<t}
\right).
\label{eq:autoregressive_action_probability}
\end{equation}
Under the controlled setting established in Section~\ref{sec:peb_definition}, the interaction history and candidate actions remain unchanged across exposure levels. Varying $k$ therefore changes only the frequency with which the target procedural pattern appears in the context, while keeping the current interaction state and the relevance of each candidate action fixed.

Repeated procedural information can enter Transformer~\citep{vaswani2017attention} computation through attention aggregation. At a generation position $t$, the attention output aggregates information from preceding context positions as:
\begin{equation}
\mathbf{o}_t
=
\sum_{j=1}^{N}
\alpha_{t,j}\mathbf{v}_j ,
\label{eq:attention_aggregation}
\end{equation}
where $N$ is the number of visible context positions when predicting $a_t$, $\alpha_{t,j}$ is the attention weight assigned to position $j$, $\mathbf{v}_j$ is its value vector, and $\mathbf{o}_t$ is the resulting attention output.

For the action-associated keyword under manipulation, increasing the exposure level $k$ introduces additional occurrences of the same procedural pattern into the model context. Each occurrence occupies a distinct context position whose representation can contribute to the attention aggregation used for subsequent action prediction. Thus, increasing $k$ increases the number of context positions carrying related procedural evidence, while leaving the interaction state and the applicability of the corresponding action unchanged. Importantly, this effect does not require any individual occurrence to receive a larger attention weight: the exposure manipulation changes the number of potential contributors to the aggregation rather than directly increasing the weight of the action occurrence.

\subsection{Reducing Procedural Exposure Bias through Visual Skill Encoding}
\label{sec:methodology_representation}
To investigate why identical procedural exposure can produce different effects under different representations, we examine how the supplied skill is encoded before contributing to action prediction under Text and Visual input. We keep the procedural content and exposure frequency consistent across the two settings, while varying how the skill is encoded and delivered to the language model.

For the \textit{Text}-based encoding, the skill is directly tokenized and inserted into the language model context. Repeating an action-associated keyword therefore introduces additional occurrences of the corresponding lexical pattern at distinct token positions. These positions directly participate in the Transformer computation described in Eq.~\ref{eq:attention_aggregation}, allowing repeated procedural evidence to contribute to action prediction through the language context.
For the \textit{Visual}-based encoding, the same procedural content is first rendered into an image. Following the standard Vision Transformer formulation~\citep{DBLP:conf/iclr/DosovitskiyB0WZ21}, the rendered image is divided into local patches, converted into patch-level representations, and processed by the model's vision encoder before being passed to the language model. Consequently, repeated procedural content is no longer introduced as repeated lexical token occurrences in the language context. Instead, each rendered occurrence is encoded through the image patches covering its spatial region and transformed by the visual encoding pathway before contributing to downstream action prediction. The two settings thus preserve the same procedural content and exposure frequency while routing the corresponding procedural evidence through different intermediate representations before reaching the language model. Further details of the matched Text and Visual skill rendering procedure are provided in Appendix~\ref{app:representation_methodology}.
\section{Experimental Methodology}
\label{sec:experimental_setup}

This section first describes the datasets, evaluation metrics, and baselines. We then present the implementation details of our experiments.

\textbf{Datasets.}
\label{sec:experimental_benchmarks}
We evaluate on three interactive benchmarks: ALFWorld~\citep{DBLP:conf/iclr/ShridharYCBTH21}, ScienceWorld~\citep{wang2022scienceworld}, and WebShop~\citep{DBLP:conf/nips/Yao0YN22}, covering household tasks, scientific interaction, and online shopping, respectively. 
ALFWorld is additionally used for controlled procedural-exposure experiments, while our ScienceWorld evaluation covers diverse scientific task categories. Detailed benchmark descriptions are provided in Appendix~\ref{app:benchmarks_metrics}.

\textbf{Evaluation Metrics.}
We report task success rate as the primary metric, together with interaction steps where applicable. Macro-average success rate is computed equally across the three benchmarks. For controlled experiments, we measure distractor preference, exposure sensitivity, and executable competing-action selection rates. Further details of benchmark-level evaluation metrics are provided in Appendix~\ref{app:evaluation_metrics}.

\textbf{Baselines.}
\label{sec:skill_representations}
We compare \textit{No Skill}, \textit{Text Skill}, and \textit{Visual Skill}. Text and Visual use identical ranked skills and procedural content, differing only in representation. Text supplies language tokens directly, whereas Visual uses rendered skill cards. An additional \textit{Rasterized} condition isolates visual encoding from structured layout.

\textbf{Implementation Details.}
\label{sec:complete_agent_setup}
We evaluate Qwen3.5-9B, Qwen3.5-27B, Gemma-4-31B-it, and Kimi-K2.5. Following SkillNet~\citep{liang2026skillnet}, Qwen3.8-Max~\citep{qwen38} ranks skills for each task. Controlled PEB experiments use 165 ALFWorld decision states with exposure levels $k\in\{0,1,2,4,8\}$, while executable-action experiments use 185 states from 64 trajectories. The skill-ranking prompt is provided in Appendix~\ref{app:skill_ranking_prompt}.

\begin{table*}[t]
\centering
\small
\setlength{\tabcolsep}{3.0pt}
\renewcommand{\arraystretch}{1.08}

\caption{Main results on ALFWorld, ScienceWorld, and WebShop under the default skill setting. \textbf{Score} denotes the average task score on WebShop, \textbf{Succ.} denotes the success rate (\%), and \textbf{Steps} denotes the average number of interaction steps. \textbf{Overall} reports the macro-average success rate across the three benchmarks. Higher scores and success rates are better ($\uparrow$). For interaction steps, shorter trajectories are preferred when task success is comparable ($\downarrow$). The best result for each metric within each backbone is highlighted in \textbf{bold}.}

\begin{tabularx}{0.94\textwidth}{@{}>{\centering\arraybackslash}p{1.55cm}>{\centering\arraybackslash}p{2.00cm}*{7}{>{\centering\arraybackslash}X}@{}}
\toprule
\multirow{2}{*}{\textbf{Setting}} & \multicolumn{1}{c}{\textbf{Overall}} & \multicolumn{2}{c}{\textbf{ALFWorld}} & \multicolumn{2}{c}{\textbf{ScienceWorld}} & \multicolumn{3}{c}{\textbf{WebShop}} \\
\cmidrule(lr){2-2}\cmidrule(lr){3-4}\cmidrule(lr){5-6}\cmidrule(lr){7-9}
& \mbox{\textbf{Macro Succ.}~$\uparrow$} & \mbox{\textbf{Succ.}~$\uparrow$} & \mbox{\textbf{Steps}~$\downarrow$} & \mbox{\textbf{Succ.}~$\uparrow$} & \mbox{\textbf{Steps}~$\downarrow$} & \mbox{\textbf{Score}~$\uparrow$} & \mbox{\textbf{Succ.}~$\uparrow$} & \mbox{\textbf{Steps}~$\downarrow$} \\
\midrule

\multicolumn{9}{c}{\textbf{Qwen3.5-9B}} \\
\hdashline
No Skill & 16.21 & 15.67 & 27.59 & 18.96 & 17.04 & 38.52 & 14.00 & 13.52 \\
Text Skill & 40.70 & 66.42 & 18.28 & 42.18 & \textbf{15.36} & 39.53 & 13.50 & 12.85 \\
\rowcolor{cyan!3} Visual Skill & \textbf{44.06} & \textbf{70.15} & \textbf{17.75} & \textbf{45.02} & 15.95 & \textbf{40.77} & \textbf{17.00} & \textbf{12.79} \\
\midrule

\multicolumn{9}{c}{\textbf{Qwen3.5-27B}} \\
\hdashline
No Skill & 41.33 & 58.96 & 19.01 & 45.02 & 15.83 & 49.27 & 20.00 & 12.10 \\
Text Skill & 53.06 & 78.36 & 14.87 & 57.82 & 14.28 & \textbf{53.93} & \textbf{23.00} & 11.96 \\
\rowcolor{cyan!3} Visual Skill & \textbf{53.94} & \textbf{80.60} & \textbf{14.66} & \textbf{59.72} & \textbf{13.80} & 51.05 & 21.50 & \textbf{9.45} \\
\midrule

\multicolumn{9}{c}{\textbf{Gemma-4-31B}} \\
\hdashline
No Skill & 41.02 & 62.69 & 18.66 & 40.76 & 16.23 & 49.66 & 19.60 & \textbf{9.47} \\
Text Skill & 57.24 & \textbf{90.30} & 12.01 & 58.29 & 14.99 & 52.66 & 23.12 & 9.92 \\
\rowcolor{cyan!3} Visual Skill & \textbf{58.21} & \textbf{90.30} & \textbf{11.78} & \textbf{59.72} & \textbf{14.60} & \textbf{52.69} & \textbf{24.62} & 9.60 \\
\midrule

\multicolumn{9}{c}{\textbf{Kimi-K2.5}} \\
\hdashline
No Skill & 49.73 & 78.36 & 16.76 & 48.82 & 16.54 & 50.64 & 22.00 & 15.11 \\
Text Skill & 56.34 & 90.30 & 12.75 & 50.71 & \textbf{15.95} & 54.88 & 28.00 & \textbf{13.35} \\
\rowcolor{cyan!3} Visual Skill & \textbf{58.64} & \textbf{91.79} & \textbf{12.41} & \textbf{52.13} & 16.06 & \textbf{56.71} & \textbf{32.00} & 13.43 \\
\bottomrule
\end{tabularx}

\label{tab:main_results}
\vspace{-10pt}
\end{table*}

\section{Results and Analysis}
\label{sec:experimental_results}

We examine how skill representation affects agent performance and exposure-dependent action prediction. We first evaluate skill representations across models and interactive environments. Controlled experiments then separate visual encoding from layout, probe repeated-signal accessibility, and assess executable action selection. Finally, layer-wise readouts and component interventions characterize exposure-dependent evidence and architecture-dependent responses.

\subsection{Visual Skills Improve Overall Agent Performance}
\label{sec:overall_evaluation}

We first test whether representation-dependent effects observed at controlled decision steps are reflected in complete agent interaction. We evaluate Qwen3.5-9B, Qwen3.5-27B, Gemma-4-31B-it, and Kimi-K2.5 on ALFWorld~\citep{DBLP:conf/iclr/ShridharYCBTH21}, ScienceWorld~\citep{wang2022scienceworld}, and WebShop~\citep{DBLP:conf/nips/Yao0YN22}, spanning different model backbones and interactive tasks.

We compare \textit{No Skill}, \textit{Text Skill}, and \textit{Visual Skill}. No Skill omits retrieved procedural guidance, while Text and Visual receive identical ranked skills and procedural content, isolating the effect of representation. Following SkillNet~\citep{liang2026skillnet}, Qwen3.8-Max~\citep{qwen38} ranks the shared skill library for each task using its goal and skill metadata. Selected skills are then supplied as text or rendered images.

Table~\ref{tab:main_results} shows that both skill representations improve macro-average success over No Skill for all four models. Visual outperforms Text in macro-average success for every backbone, although their relative performance varies by environment. These results do not establish whether reduced procedural exposure bias fully explains the performance advantage. We therefore examine representation-dependent action prediction through controlled experiments. Further task-level results are provided in Appendix~\ref{app:complete_agent_evaluation}.

\subsection{Why Visual Skills Reduce Procedural Exposure Bias}
\label{sec:representation_analysis}

We investigate the weaker influence of procedural exposure under Visual, following the encoding pathways in Section~\ref{sec:methodology_representation}. We disentangle visual encoding from layout, probe repeated-signal accessibility and executable action selection, and examine layer-wise evidence and component-level interventions. Additional protocols and results are provided in Appendix~\ref{app:representation_experiments}.

\paragraph{Visual Encoding Attenuates Procedural Exposure Bias.}
\label{sec:representation_intervention}

\begin{figure*}[t]
    \centering
    \begin{subfigure}[t]{0.32\textwidth}
        \centering
        \includegraphics[width=\linewidth]{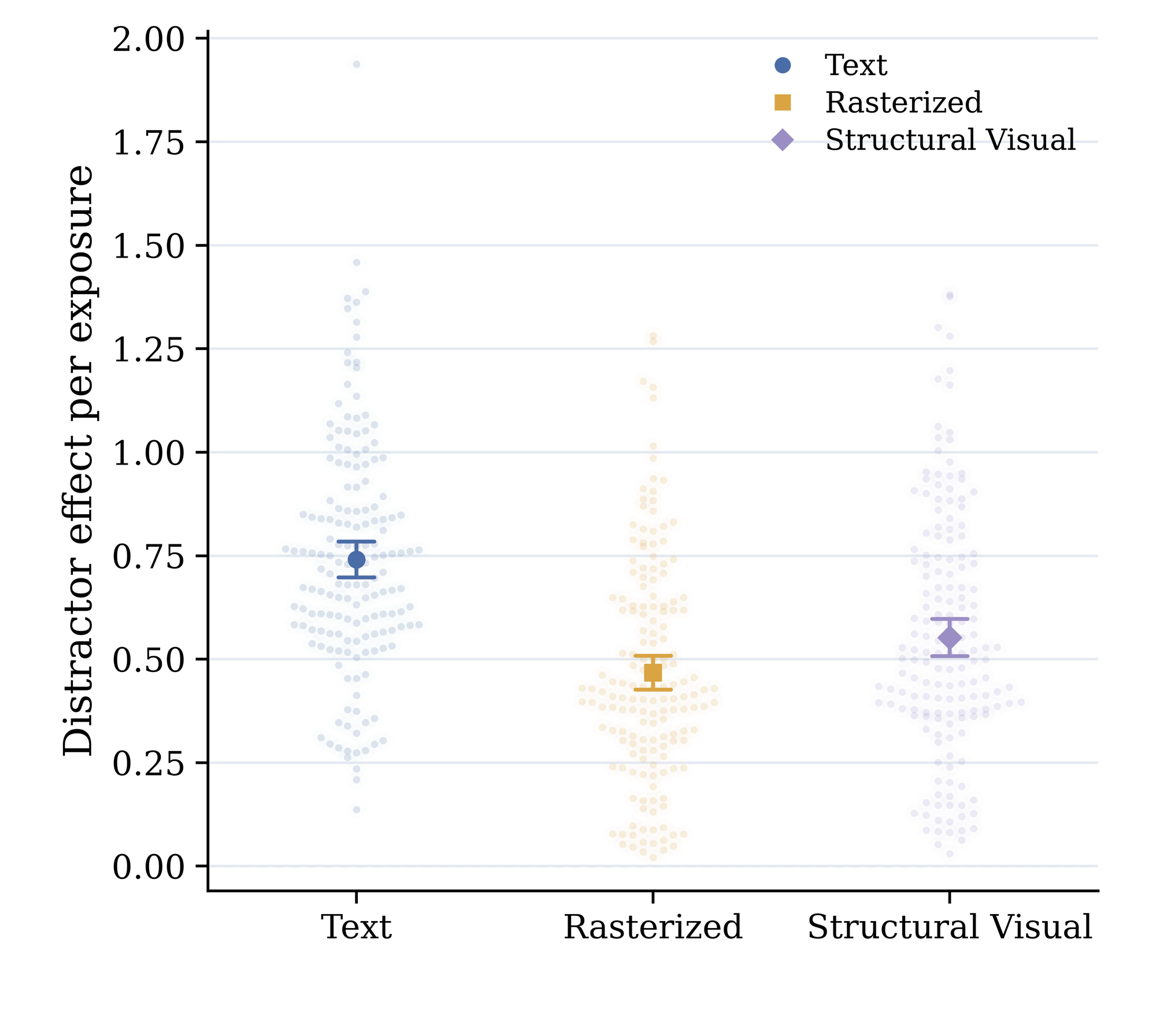}
        \caption{Encoding Pathway Intervention.}
        \label{fig:ri_9b}
    \end{subfigure}
    \hfill
    \begin{subfigure}[t]{0.32\textwidth}
        \centering
        \includegraphics[width=\linewidth]{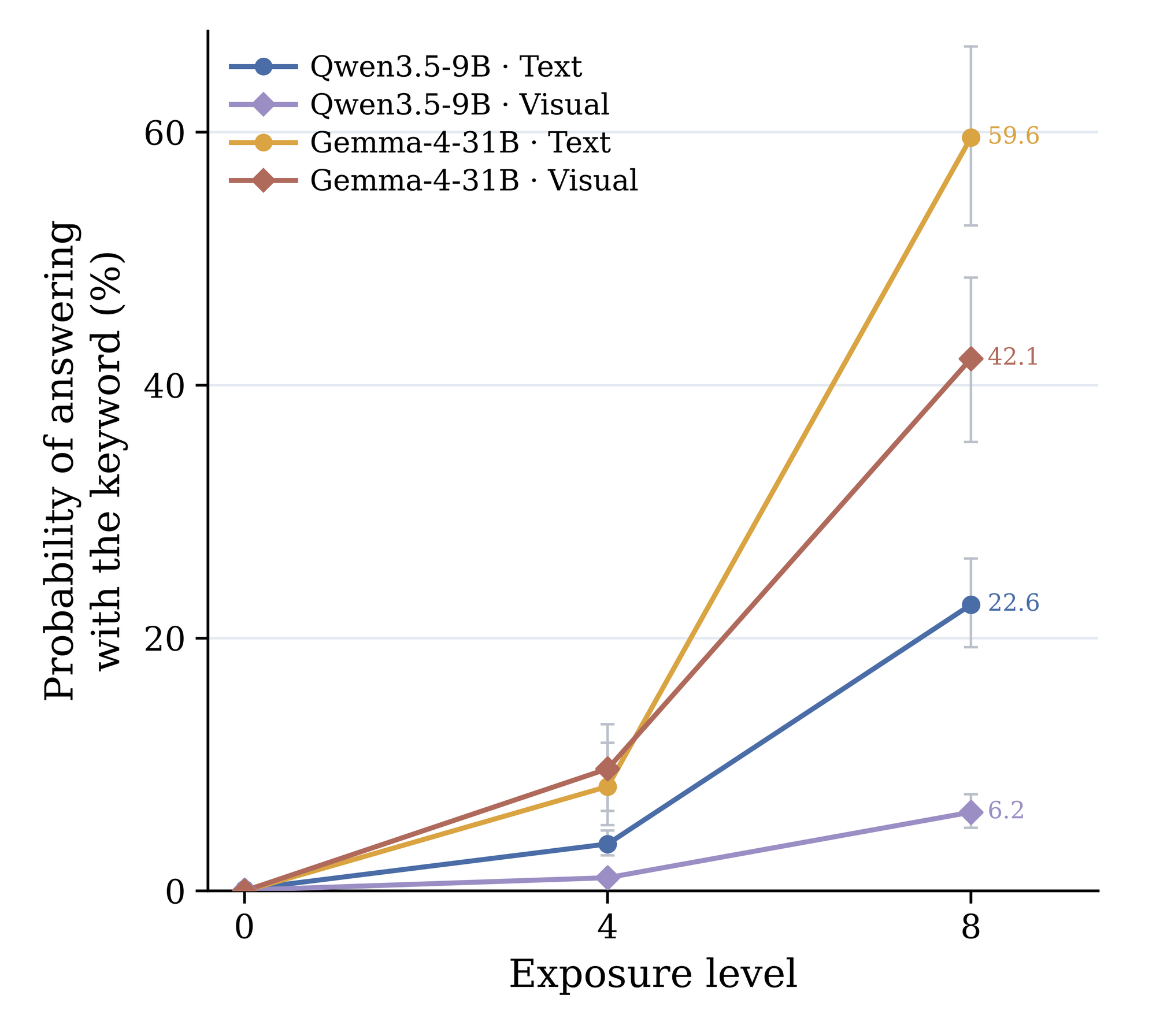}
        \caption{Repeated Signal Accessibility.}
        \label{fig:target_word_readout}
    \end{subfigure}
    \hfill
    \begin{subfigure}[t]{0.32\textwidth}
        \centering
        \includegraphics[width=\linewidth]{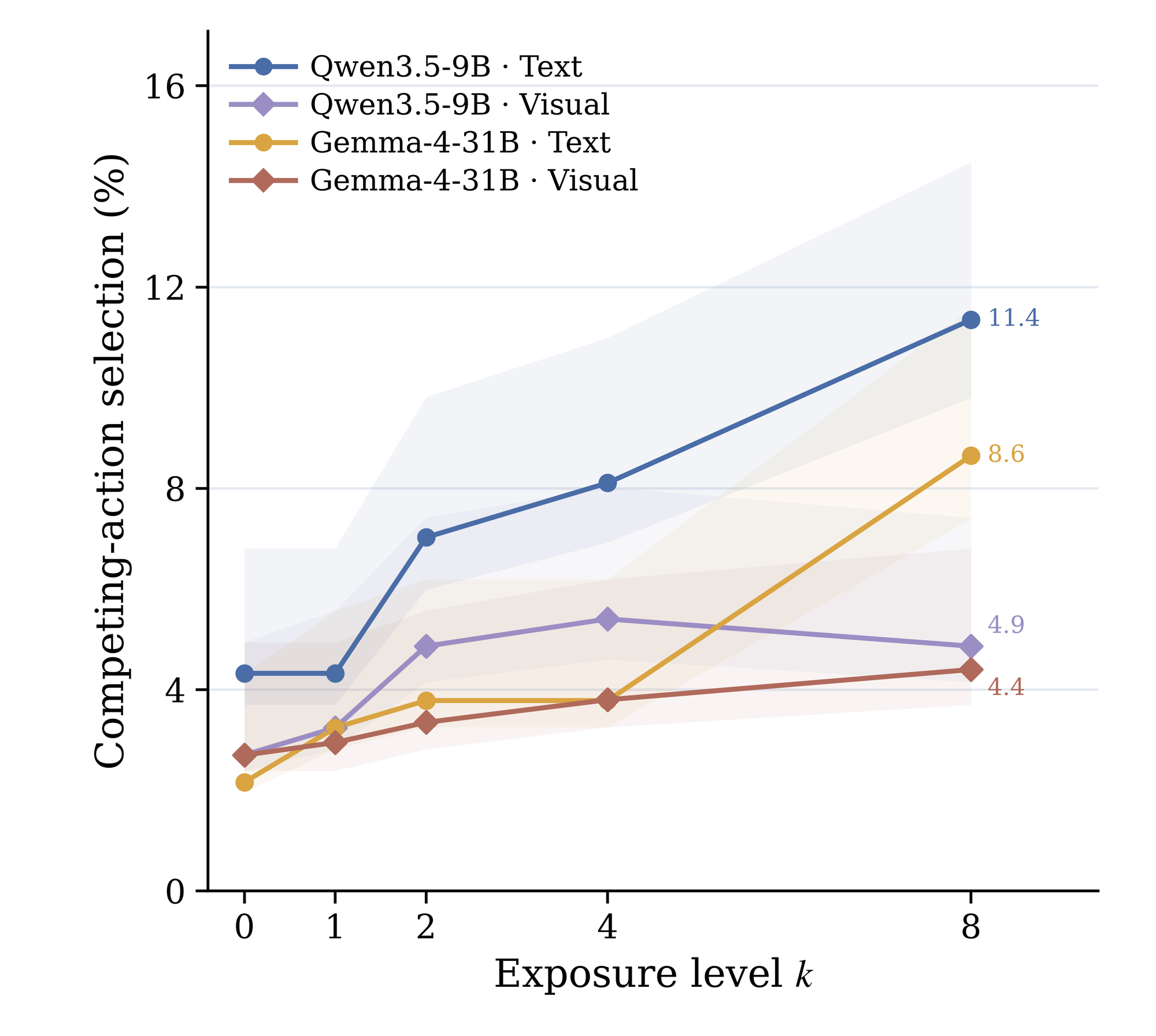}
        \caption{Executable Action Selection.}
        \label{fig:behavioral_realization}
    \end{subfigure}
    \caption{
    Representation-Dependent Effects of Procedural Exposure.
    Figure~\ref{fig:ri_9b} compares exposure sensitivity under Text, Rasterized, and Structural Visual representations.
    Figure~\ref{fig:target_word_readout} reports exposure-dependent target-word probability under Text and Visual encoding.
    Figure~\ref{fig:behavioral_realization} reports the greedy selection rate of an executable competing action as procedural exposure increases.
    }
    \label{fig:representation_analysis}
    \vspace{-12pt}
\end{figure*}

To separate visual encoding from structured layout, we compare three representations of identical procedural content. \textit{Text} supplies language tokens directly; \textit{Rasterized} preserves the wording and ordering in a plain monospaced image; and \textit{Structural Visual} uses a structured skill-card format based on the main-experiment design, separately implemented for the controlled probes.

Figure~\ref{fig:ri_9b} shows that Qwen3.5-9B exhibits substantially lower exposure sensitivity under Rasterized than under Text, while Structural Visual shows comparable attenuation. This supports visual encoding, rather than layout alone, as the primary source of attenuation. Qwen3.5-27B and Gemma-4-31B-it show the same qualitative pattern, as detailed in Appendix~\ref{app:rasterized_interference}.

We next test whether weaker exposure sensitivity under Visual can be explained by reduced access to repeated procedural content. Rather than scoring candidate actions, we prompt the model to identify the most frequent object and score the target word at a fixed decoding position for $k\in\{0,4,8\}$. Figure~\ref{fig:target_word_readout} shows that target-word probability rises with repetition under both Text and Visual for Qwen3.5-9B and Gemma-4-31B-it. Repetition thus continues to affect target-word prediction after visual encoding, despite its weaker influence on distractor-action prediction. This argues against an explanation based solely on information loss, but does not establish equal accessibility across representations. Readout controls and scoring details are provided in Appendix~\ref{app:signal_readout}.

Finally, we test whether exposure-dependent preference shifts affect generated actions. In a separate controlled setting, the state-inapplicable distractor is replaced by an admissible, task-relevant competing action, and we measure its greedy selection rate across exposure levels. Figure~\ref{fig:behavioral_realization} shows that selection increases more strongly under Text than Visual. Thus, the representation-dependent effect extends beyond fixed-candidate scoring to executable actions. Additional action-selection analyses and paired rollout results are provided in Appendices~\ref{app:action_selection_details} and~\ref{app:paired_rollouts}, respectively.

\begin{wrapfigure}[16]{r}{0.48\columnwidth}
    \centering
    \vspace{0em}
    \begin{subfigure}[t]{0.48\linewidth}
        \centering
        \includegraphics[width=\linewidth]{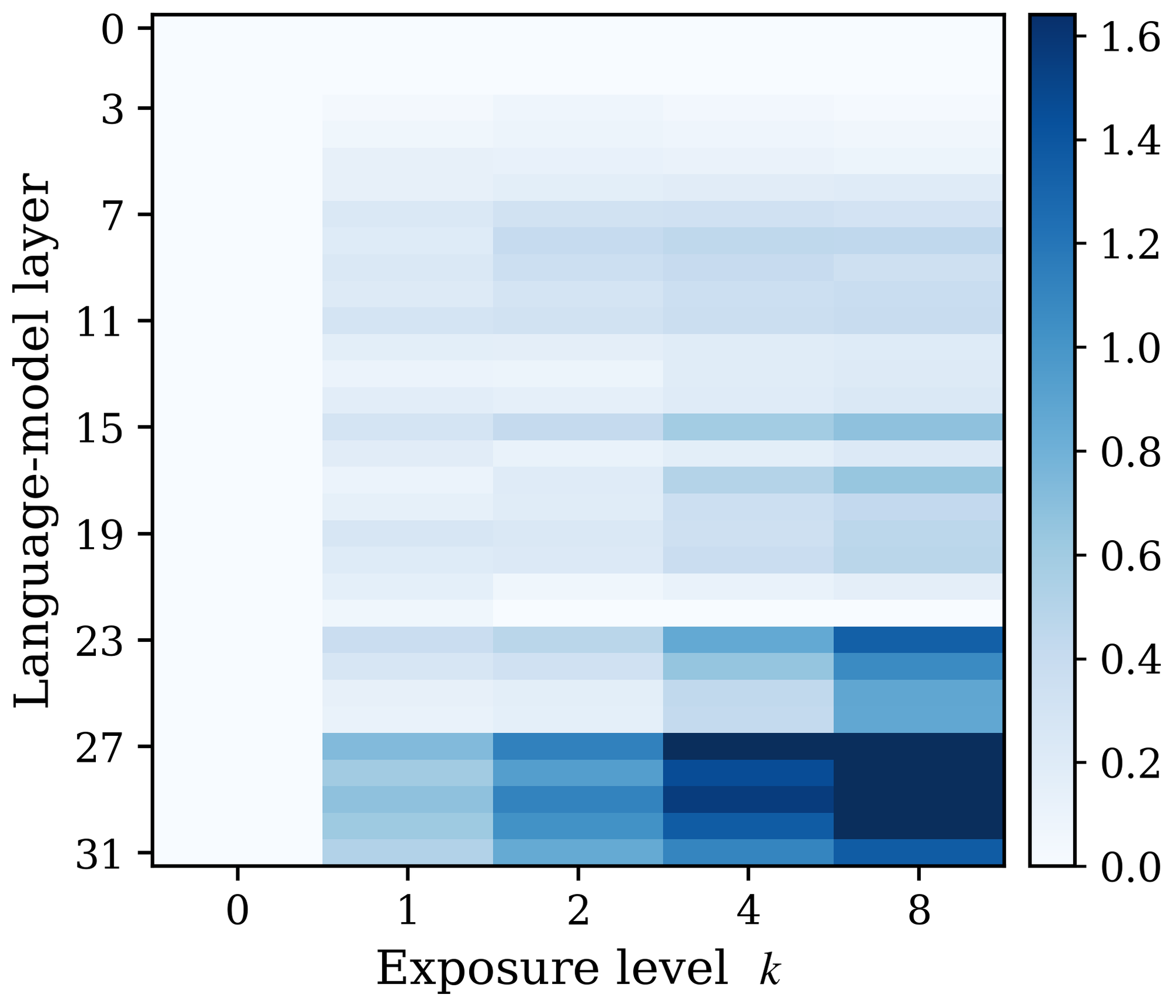}
        \caption{Text.}
        \label{fig:layerwise_qwen_text}
    \end{subfigure}
    \hfill
    \begin{subfigure}[t]{0.48\linewidth}
        \centering
        \includegraphics[width=\linewidth]{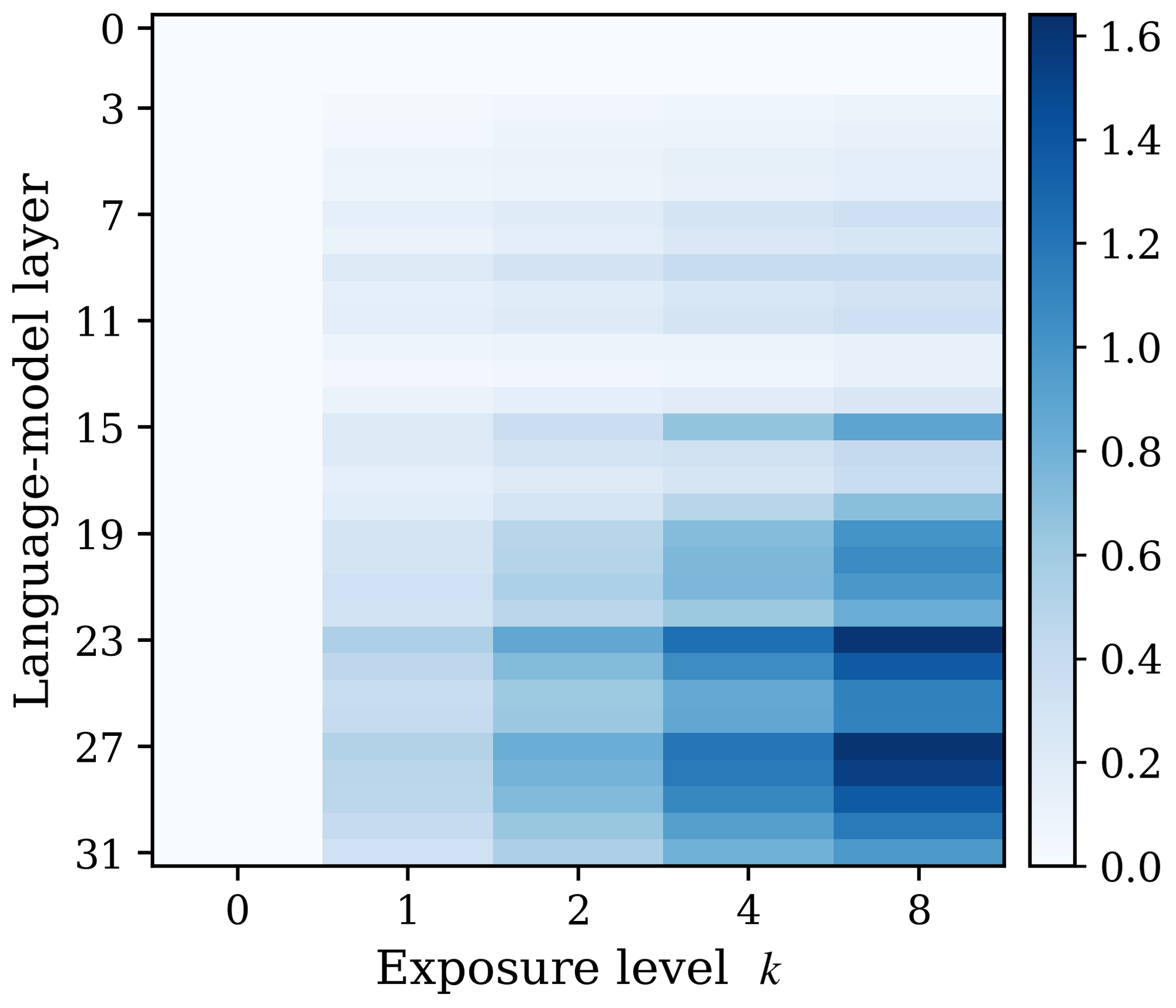}
        \caption{Visual.}
        \label{fig:layerwise_qwen_visual}
    \end{subfigure}
    \caption{
    Layer-Wise Readout of Exposure-Dependent Action Evidence in Qwen3.5-9B.
    Values show the mean change in length-normalized distractor action log-probability relative to the $k=0$ baseline.
    }
    \label{fig:layerwise_exposure}
    \vspace{0em}
\end{wrapfigure}

\paragraph{Exposure-Dependent Action Evidence Develops Across Model Depth.}
\label{sec:internal_development}

The preceding experiments show that repetition continues to influence target-word prediction under Visual, while its effect on distractor-action prediction is attenuated. We next examine how this representation-dependent difference is expressed across model depth. At each layer, we apply the final normalization and language-model head to intermediate representations at the candidate-action positions, obtaining a layer-wise readout of exposure-dependent action evidence.

Figure~\ref{fig:layerwise_exposure} shows that exposure-dependent distractor support is weakly expressed in early-layer readouts and becomes more pronounced through the middle and later layers. The representations diverge at greater depth: Text exhibits a stronger increase in exposure-dependent distractor support, whereas Visual shows a substantially weaker increase. This observation is consistent with the different representation pathways described in Section~\ref{sec:methodology_representation}, while the readout itself does not identify the internal computations responsible for the difference. Additional layer-wise results for Qwen3.5-27B and Gemma-4-31B-it are provided in Appendix~\ref{app:layerwise_readout}.

\paragraph{Component Interventions Reveal Architecture-Dependent Responses.}
\label{sec:component_intervention}

\begin{wrapfigure}[16]{r}{0.48\columnwidth}
    \centering
    \vspace{6pt}
    \captionsetup{skip=2pt}
    \begin{subfigure}[t]{0.48\linewidth}
        \centering
        \includegraphics[width=\linewidth]{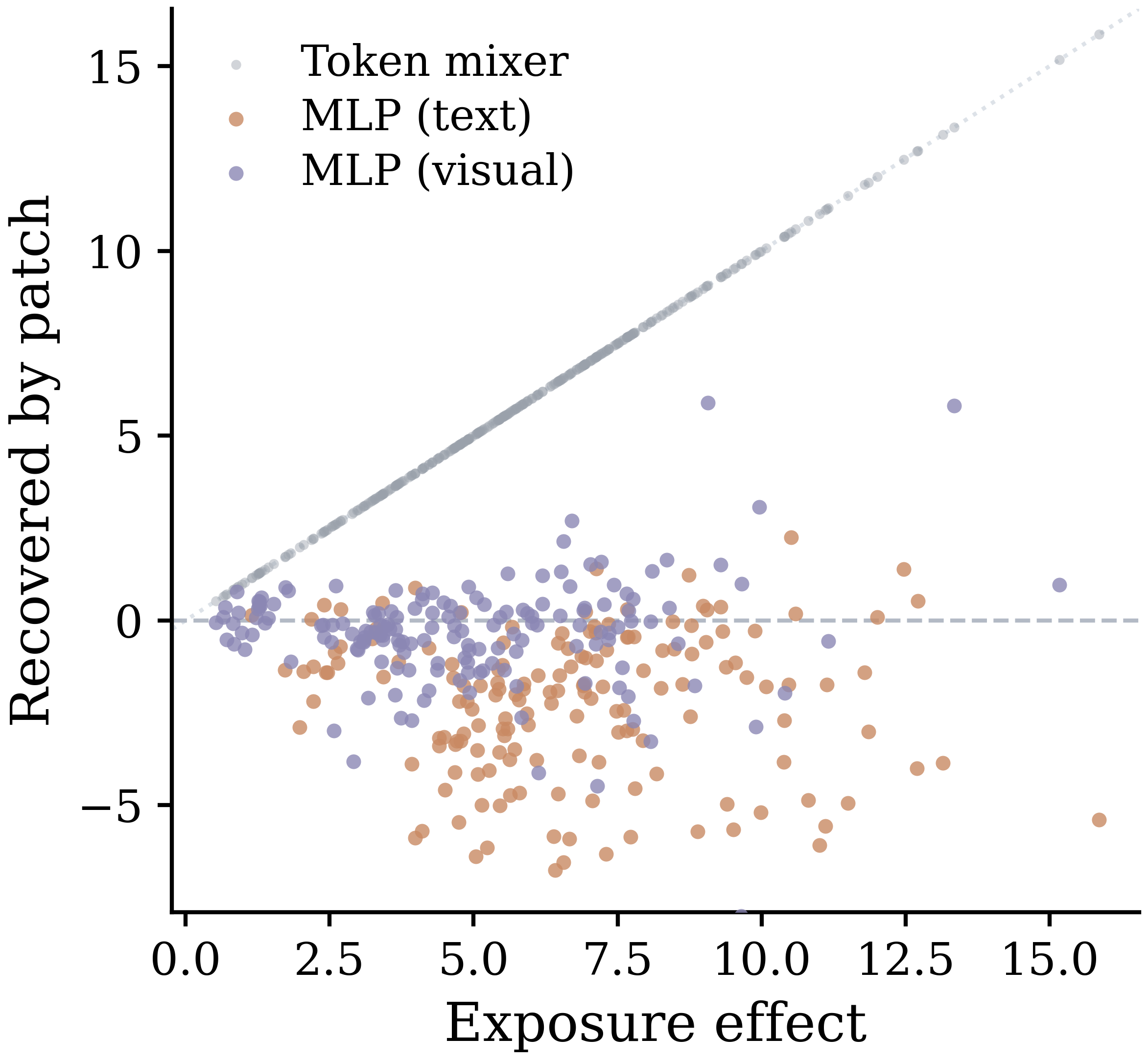}
        \caption{Qwen3.5-9B.}
        \label{fig:component_patching_qwen}
    \end{subfigure}
    \hfill
    \begin{subfigure}[t]{0.48\linewidth}
        \centering
        \includegraphics[width=\linewidth]{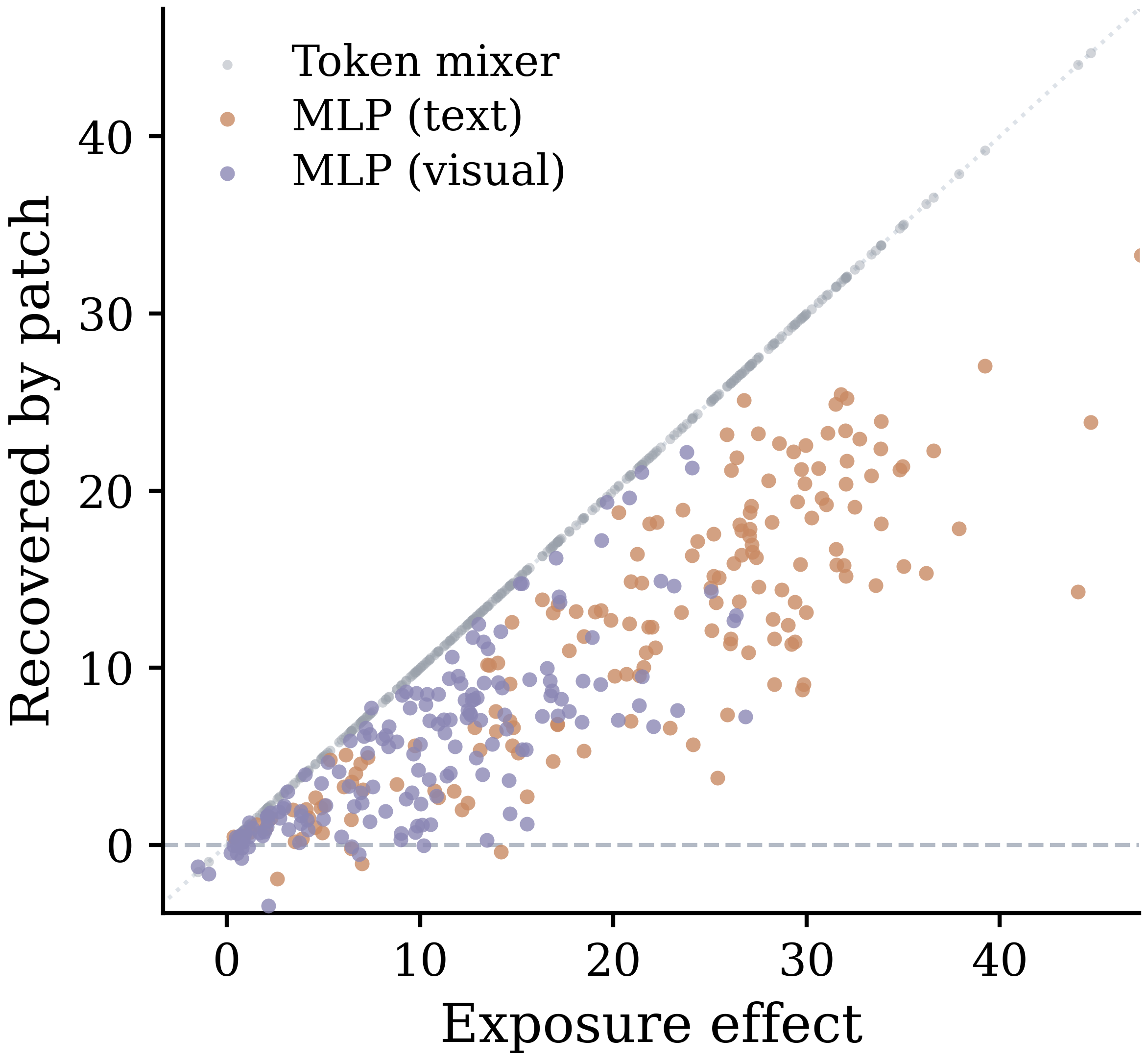}
        \caption{Gemma-4-31B-it.}
        \label{fig:component_patching_gemma}
    \end{subfigure}
    \vspace{0pt}
    \caption{
    Component Intervention Under Controlled Procedural Exposure.
    The diagonal denotes full recovery of the exposure-induced change, and the horizontal dashed line denotes zero recovery.
    }
    \label{fig:component_patching}
\end{wrapfigure}

Although the layer-wise readout reveals where exposure-dependent action evidence becomes decodable, it does not establish the contribution of individual components to the final score change. We therefore conduct matched activation-patching interventions between zero-exposure ($k=0$) and high-exposure ($k=8$) runs. During the $k=0$ pass, we replace a selected component output at the distractor-action positions with the corresponding cached activation from the $k=8$ run and measure the resulting change in the distractor score. Figure~\ref{fig:component_patching} compares this intervention-induced change with the original exposure-induced change. Equality between the two corresponds to full recovery, while a zero intervention-induced change indicates no recovery under the tested replacement.

The interventions reveal different component-level responses across architectures. In Qwen3.5-9B, MLP-output patching produces score changes opposite in direction to the original exposure effect under both representations, as shown in Figure~\ref{fig:component_patching_qwen}. In contrast, the same intervention in Gemma-4-31B-it produces partial recovery of the exposure-induced score increase, as shown in Figure~\ref{fig:component_patching_gemma}. Qwen3.5-27B exhibits a pattern similar to Qwen3.5-9B, with additional results provided in Appendix~\ref{app:component_patching}. These results indicate that the response to a matched component intervention is architecture-dependent, even when the models show a similar overall reduction in exposure sensitivity under Visual. The interventions thus provide complementary evidence about the internal expression of procedural exposure effects without attributing Visual's attenuation of PEB to a single downstream component.

\section{Related Work}
\label{sec:related_work}

Recent work on skill-augmented language-model agents has studied the broader lifecycle of reusable procedural knowledge, including skill acquisition, organization, retrieval, and deployment~\citep{zhou2026comprehensive}. Skill acquisition and evolution methods construct, refine, and reuse procedural knowledge from prior experience~\citep{wang2026skillx}, while recent benchmarks examine whether access to such skills leads to reliable improvements in downstream agent performance~\citep{zhong2026skilllearnbench}. As skill libraries continue to grow, retrieval-based approaches increasingly focus on selecting task-relevant procedures from large collections~\citep{zheng2026skillrouter,wang2026skill}, and injection-oriented pipelines further determine which retrieved information is ultimately supplied to the model~\citep{li2026skillsinjector}. Together, these studies have substantially improved how procedural knowledge is acquired and made available to language-model agents, while also highlighting the importance of how supplied skills are ultimately utilized during state-conditioned decision making.

Language models are also known to be sensitive to statistical regularities and repeated patterns in contextual input~\citep{cheng2025llms}. In in-context learning, variations in demonstration composition and placement can induce majority-label and positional biases, causing predictions to reflect contextual statistics or presentation patterns rather than the intended input--label relation~\citep{gupta2023robust,cobbina2025show}. Shortcut-learning studies show that models can rely on repeated lexical patterns or surface cues rather than the intended semantic relation~\citep{doan2025understanding,tasawong2025shortcut}. Explicit repetition can also systematically alter language-model reasoning behavior~\citep{yu2025unleashing}. These effects arise in a broader linguistic setting where token and word frequencies are highly non-uniform, as classically characterized by Zipf's law~\citep{piantadosi2014zipf,dkebowski2026zipf}. Related exposure effects have also been studied in recommender systems, where frequently exposed or popular items can receive preference disproportionate to their relevance~\citep{he2025relieving,carnovalini2025popularity}. These lines of work collectively demonstrate that frequency, repetition, and contextual statistics can systematically shape model preference, although their effect within supplied procedural knowledge remains less explored.

The representation through which information is supplied can further affect how that information is processed and utilized~\citep{van2026same}. Prior work has shown that textual content can be rendered into visual form while preserving its underlying information~\citep{li2025text}, while visually rendered inputs are processed through a distinct visual encoding pathway before being integrated into language-model computation~\citep{wang2026render}. Visual representations have also been extensively studied in agents operating over structured interfaces and multimodal observations~\citep{liu2025visualagentbench,xu2024aguvis,hasegawa2025tama}. In parallel, mechanistic studies have developed circuit-level accounts of Transformer computation and identified recurring mechanisms associated with in-context learning~\citep{elhage2021mathematical,olsson2022context}. More recent work has extended these analyses to layer-wise dynamics and broader patterns of information flow within Transformer models~\citep{atad-etal-2026-tensorlens}. Building on these research directions, we study how repeated procedural information within an already supplied skill influences action prediction, and how this effect changes when the same procedural content is represented through textual or visual encoding pathways.
\section{Conclusion}
\label{sec:conclusion}

This paper investigates how procedural exposure affects skill utilization in language-model agents. We identify and formalize \textbf{Procedural Exposure Bias (PEB)}, showing that repeated exposure to an action-associated procedural pattern can systematically shift action preference even when the interaction state and action relevance remain unchanged. We further show that this effect is strongly representation-dependent: visual encoding substantially attenuates PEB while keeping the repeated procedural signal recoverable, and internal analyses reveal architecture-dependent pathways through which exposure-related action evidence develops. Finally, controlled generation and complete-agent evaluations show that these representation differences extend to executable decisions and persist across ALFWorld, ScienceWorld, and WebShop. These findings identify skill representation as an important factor in how language-model agents utilize externally supplied procedural knowledge.

\section*{AI Use Statement}
Generative AI tools were used to assist with language polishing and improving the clarity of the manuscript. All research ideas, experimental design, implementation, analysis, and conclusions were developed and verified by the authors, who take full responsibility for the content of the paper.

\bibliography{iclr2027_conference}
\bibliographystyle{iclr2027_conference}

\clearpage
\appendix

\section{Appendix}
\label{sec:appendix}

This appendix follows the organization of the main paper. We first provide the controlled construction and supplementary evidence for procedural exposure bias, followed by details of Text and Visual skill encoding, the shared experimental setup, and additional results for complete-agent evaluation and representation-dependent analyses.

\subsection{Details of Procedural Exposure Bias Probing}
\label{app:peb_details}

This section supplements Section~\ref{sec:background} with decision-state construction, exposure-controlled skill variants, candidate-action scoring, keyword-replacement controls, and additional-model results.

\subsubsection{Decision States, Distractors, and Procedural Keywords}
\label{app:peb_states}

We extract 165 intermediate decision steps from successful ALFWorld trajectories. Each trajectory is replayed in the environment simulator to reconstruct the observation history and admissible action set at the selected step. Procedural guidance supplied during the source run is removed from the reconstructed history before constructing the probe.

For each decision step $i$, let $a_i^{C}$ denote the action executed in the reference trajectory. We introduce a distractor action $a_i^{D}$ that is relevant to the overall task and refers to an existing environment entity but is excluded from the current admissible action set. The distractor is therefore task-relevant but state-inapplicable. Its verb is required to differ from that of the reference action.

Each distractor is associated with a target keyword $w(a_i^{D})$. Consistent with Section~\ref{sec:peb_definition}, the target is an object-related noun associated with the action; for example, \texttt{microwave} for \texttt{open microwave}. Occurrences are counted using whole-word matching, independently of the model's tokenization. The target keyword must occur in the full 37-skill corpus, but is not required to appear in the five selected skills, and must not appear in the task query or the reference action. Keyword frequency serves as the operational measure of action-associated exposure in these probes.

\subsubsection{Exposure-Controlled Skill Construction}
\label{app:peb_construction}

We construct skill variants with target exposure levels $k\in\{0,1,2,4,8\}$ while keeping the interaction state and action applicability fixed. Exposure is counted within the supplied skill, rather than across the full interaction history.

Before constructing the exposure variants, native occurrences of the target keyword throughout the descriptions and bodies of the five supplied skills are replaced using a fixed mapping of single-word neutral substitutes, while skill names remain unchanged. Fixed insertion slots are distributed across the skill document. At exposure level $k$, exactly $k$ slots contain target-bearing procedural phrases, and the remaining slots contain filler phrases without the target keyword. Target-bearing and filler phrases are matched in word count. Apart from this keyword replacement, content outside the insertion slots remains unchanged, and automated checks verify that the total whole-word occurrence count in the supplied skill equals $k$.

The total word count is held constant across exposure levels for each decision state. This word-count constraint does not imply equal token counts under every model's tokenizer.

\subsubsection{Action Scoring and Distractor Preference}
\label{app:peb_scoring}

Let $\mathcal{A}_i^{\mathrm{adm}}$ denote the admissible action set at decision step $i$. We exclude \texttt{look}, \texttt{inventory}, \texttt{help}, \texttt{wait}, and \texttt{examine} actions, and retain at most 16 admissible candidates, denoted by $\mathcal{A}_i^{\mathrm{sel}}$. The scored candidate set is
\begin{equation}
\mathcal{A}_i
=
\mathcal{A}_i^{\mathrm{sel}}
\cup
\{a_i^{D}\}.
\label{eq:peb_candidates}
\end{equation}
This set is fixed across exposure levels and skill representations for each state. Including the inadmissible distractor permits its model support to be measured without executing it in the environment.

Candidates are scored using teacher forcing with a fixed scoring prefix. We use the length-normalized token log-probability in Equation~\ref{eq:action_score}. To make the representation explicit, we write this score as $S_i(a;k,r)$, where $r$ denotes the skill representation and $x_i^{(k,r)}$ denotes the corresponding complete input. Candidate scores are normalized as
\begin{equation}
P_i(a;k,r)
=
\frac{\exp\!\left(S_i(a;k,r)\right)}
{\sum_{a'\in\mathcal{A}_i}\exp\!\left(S_i(a';k,r)\right)}.
\label{eq:peb_candidate_probability}
\end{equation}
Distractor preference is $P_i(a_i^{D};k,r)$. This quantity is a normalized preference over the fixed candidate set, not the probability that unrestricted generation produces the distractor action.

Figure~\ref{fig:procedural_exposure_bias_a} reports mean distractor preference across exposure levels. For Figure~\ref{fig:procedural_exposure_bias_b}, preference is instead computed by normalizing summed action-token log-probabilities $\widetilde{S}_i(a;k,r)=|a|S_i(a;k,r)$ across the same candidate set, yielding $\widetilde{P}_i(a;k,r)$. The within-state endpoint change is
\begin{equation}
\Delta \widetilde{P}_{i,r}^{D}
=
\widetilde{P}_i(a_i^{D};8,r)
-
\widetilde{P}_i(a_i^{D};0,r).
\label{eq:peb_delta_preference}
\end{equation}
The distribution of these paired changes characterizes variation in the exposure effect across decision states.

\subsubsection{Keyword-Replacement Control}
\label{app:peb_replacement}

At $k=8$, we compare three conditions to examine whether distractor preference depends on the target keyword rather than lexical replacement in general.

The \emph{original} condition retains the exposure-controlled skill. The \emph{target-replacement} condition replaces all occurrences of the distractor keyword with a neutral alternative while preserving word count and syntactic structure. The \emph{frequency-matched control} condition replaces an unrelated word with the same occurrence frequency while leaving the target keyword intact. Control words are selected to avoid semantic overlap with the distractor action, reference action, task query, and operational verbs.

Figure~\ref{fig:procedural_exposure_bias_c} uses the normalized preference metric defined above. The original and target-replacement conditions include all 165 decision states. A qualifying frequency-matched control word is available for 128 states, and the frequency-matched control statistics use this subset. These sample counts differ; comparisons of aggregate values across the full set and the subset are therefore not fully paired.

\subsubsection{Additional Models}
\label{app:peb_additional_models}

\begin{figure*}[t]
    \centering

    \begin{subfigure}[t]{0.32\textwidth}
        \centering
        \vspace{0pt}
        \begin{minipage}[t][4.25cm][t]{\linewidth}
            \centering
            \includegraphics[width=\linewidth]{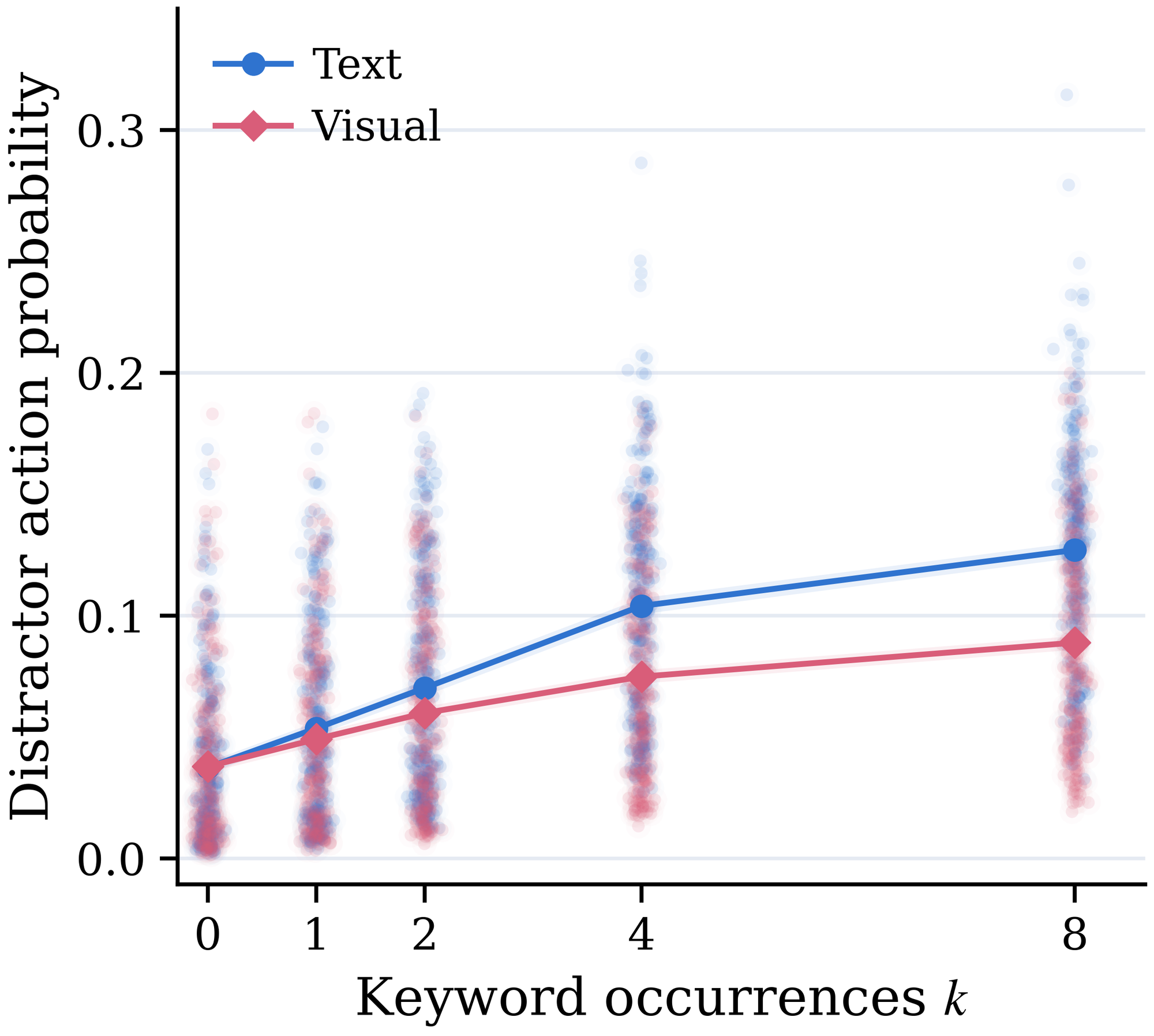}
        \end{minipage}
        \caption{Exposure-Dependent Distractor Preference.}
        \label{fig:procedural_exposure_bias_a_27}
    \end{subfigure}
    \hfill
    \begin{subfigure}[t]{0.32\textwidth}
        \centering
        \vspace{0pt}
        \begin{minipage}[t][4.25cm][t]{\linewidth}
            \centering
            \includegraphics[width=\linewidth]{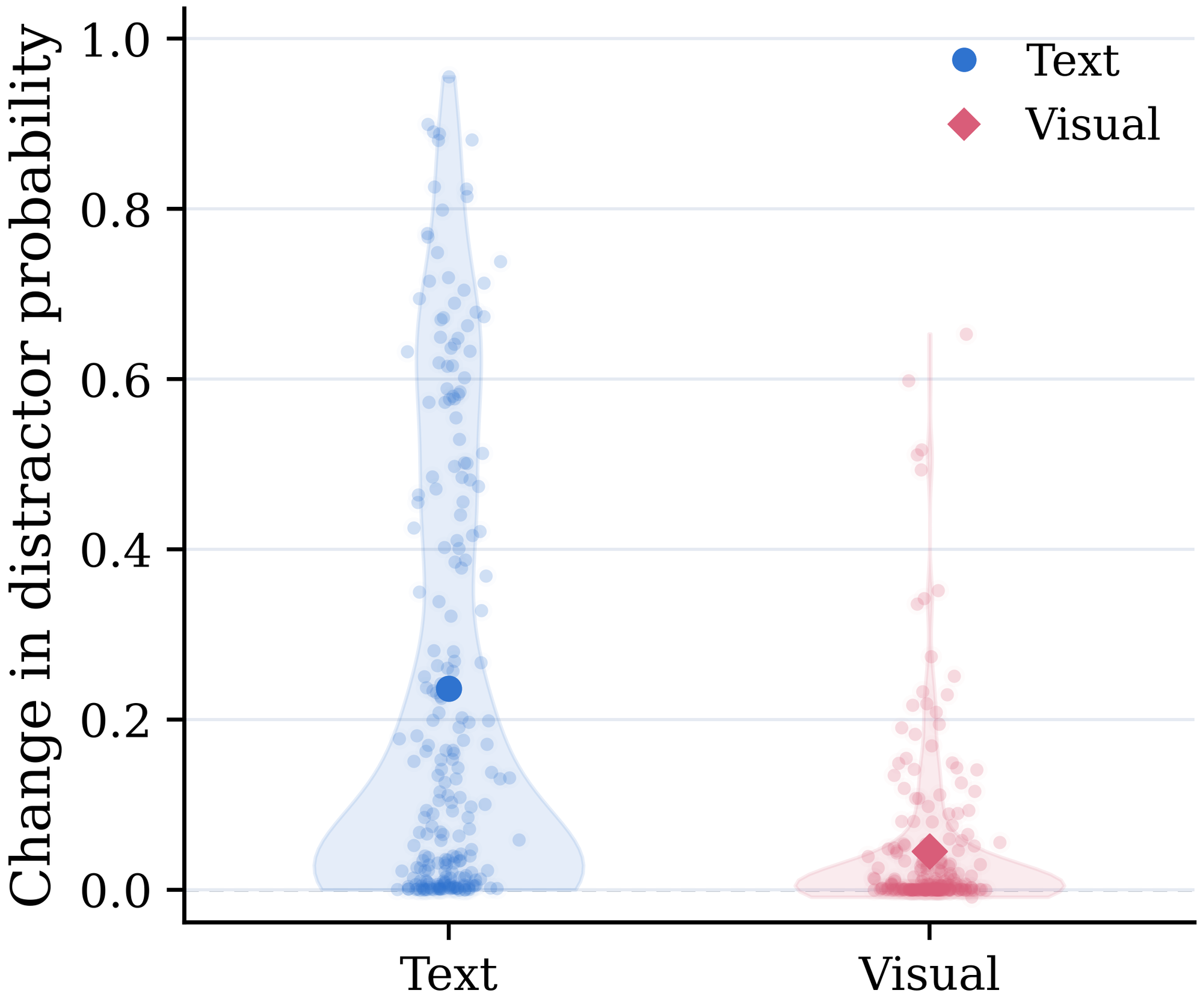}
        \end{minipage}
        \caption{Paired Changes in Distractor Preference.}
        \label{fig:procedural_exposure_bias_b_27}
    \end{subfigure}
    \hfill
    \begin{subfigure}[t]{0.32\textwidth}
        \centering
        \vspace{0pt}
        \begin{minipage}[t][4.25cm][t]{\linewidth}
            \centering
            \includegraphics[width=\linewidth]{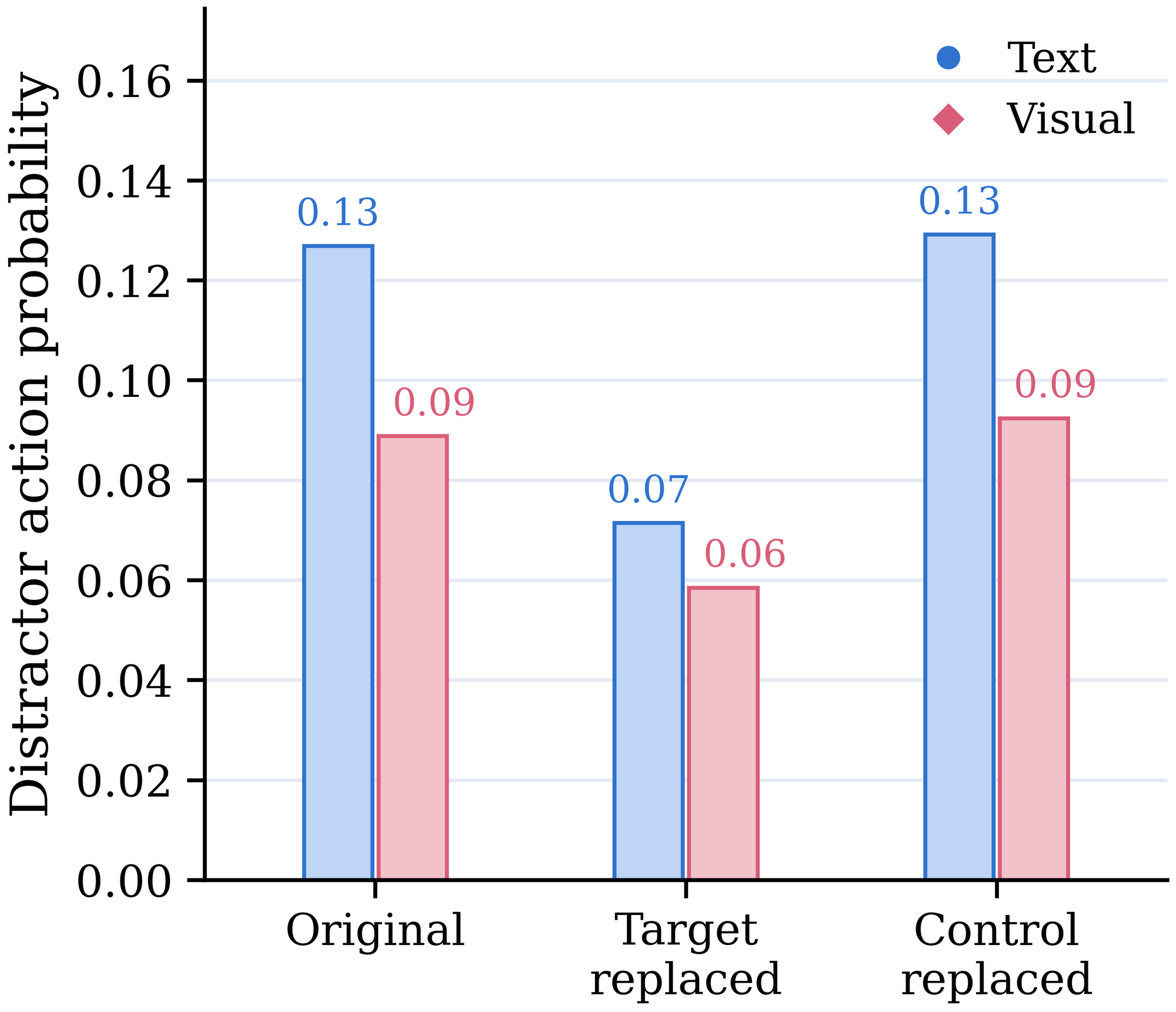}
        \end{minipage}
        \caption{Keyword-Replacement Controls.}
        \label{fig:procedural_exposure_bias_c_27}
    \end{subfigure}

    \caption{
        Procedural Exposure Bias in Qwen3.5-27B Under Text and Visual Skill Representations.
        Figure~\ref{fig:procedural_exposure_bias_a_27}: Distractor Preference Across Exposure Levels.
        Figure~\ref{fig:procedural_exposure_bias_b_27}: Paired Preference Changes From $k=0$ to $k=8$.
        Figure~\ref{fig:procedural_exposure_bias_c_27}: Keyword-Replacement Controls.
    }
    \label{fig:procedural_exposure_bias_27}
\end{figure*}

\begin{figure*}[t]
    \centering

    \begin{subfigure}[t]{0.32\textwidth}
        \centering
        \vspace{0pt}
        \begin{minipage}[t][4.25cm][t]{\linewidth}
            \centering
            \includegraphics[width=\linewidth]{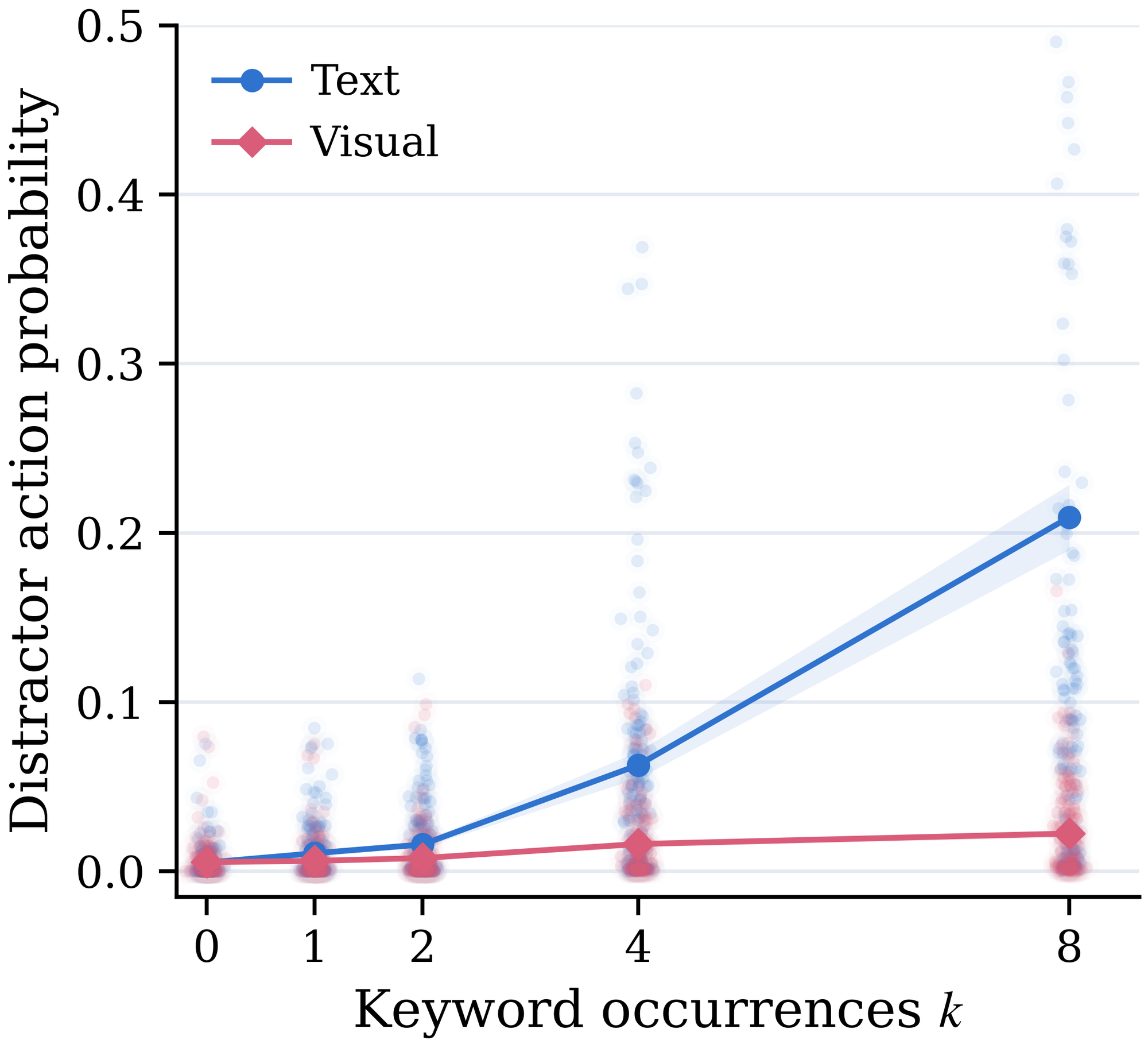}
        \end{minipage}
        \caption{Exposure-Dependent Distractor Preference.}
        \label{fig:procedural_exposure_bias_a_31}
    \end{subfigure}
    \hfill
    \begin{subfigure}[t]{0.32\textwidth}
        \centering
        \vspace{0pt}
        \begin{minipage}[t][4.25cm][t]{\linewidth}
            \centering
            \includegraphics[width=\linewidth]{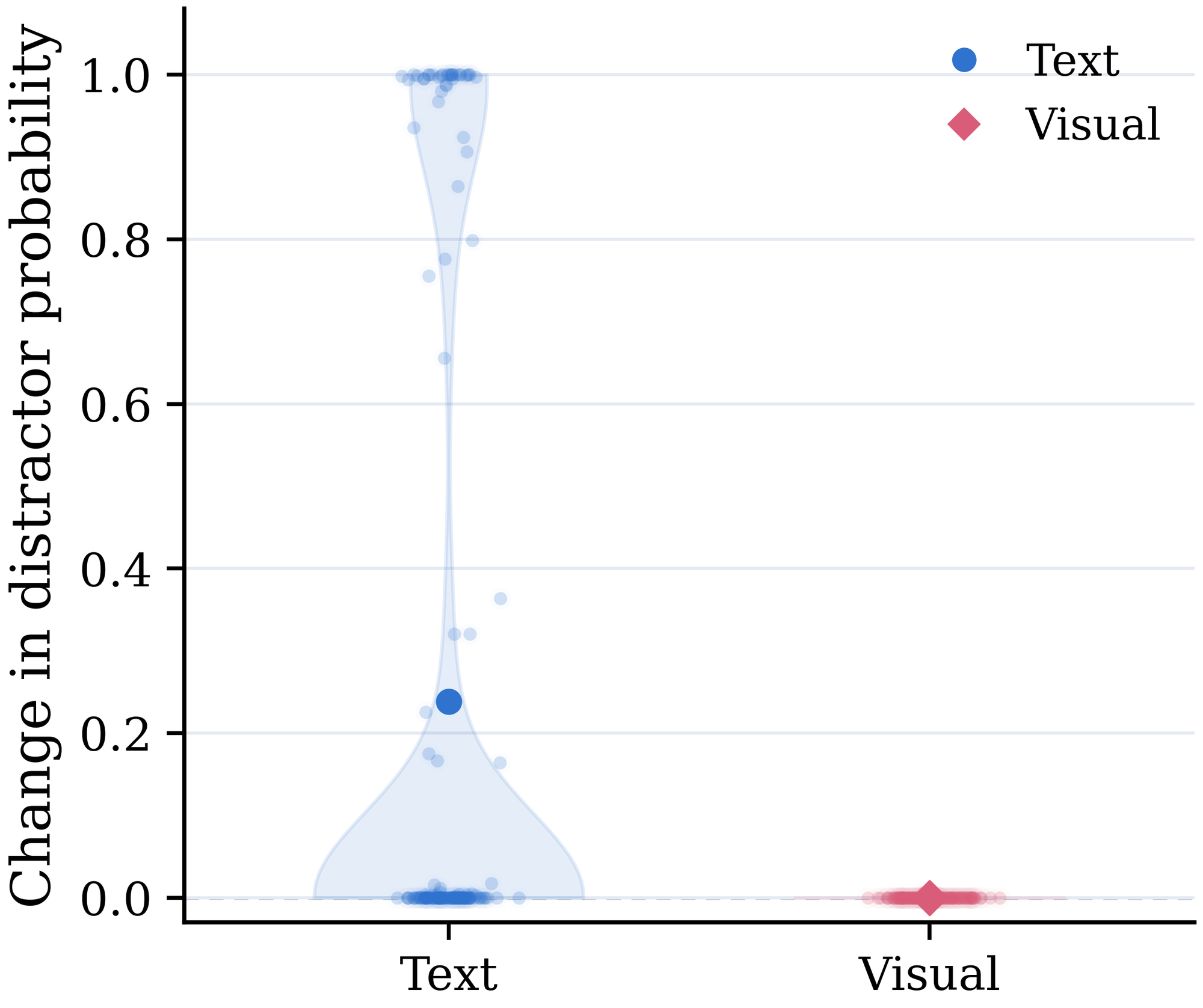}
        \end{minipage}
        \caption{Paired Changes in Distractor Preference.}
        \label{fig:procedural_exposure_bias_b_31}
    \end{subfigure}
    \hfill
    \begin{subfigure}[t]{0.32\textwidth}
        \centering
        \vspace{0pt}
        \begin{minipage}[t][4.25cm][t]{\linewidth}
            \centering
            \includegraphics[width=\linewidth]{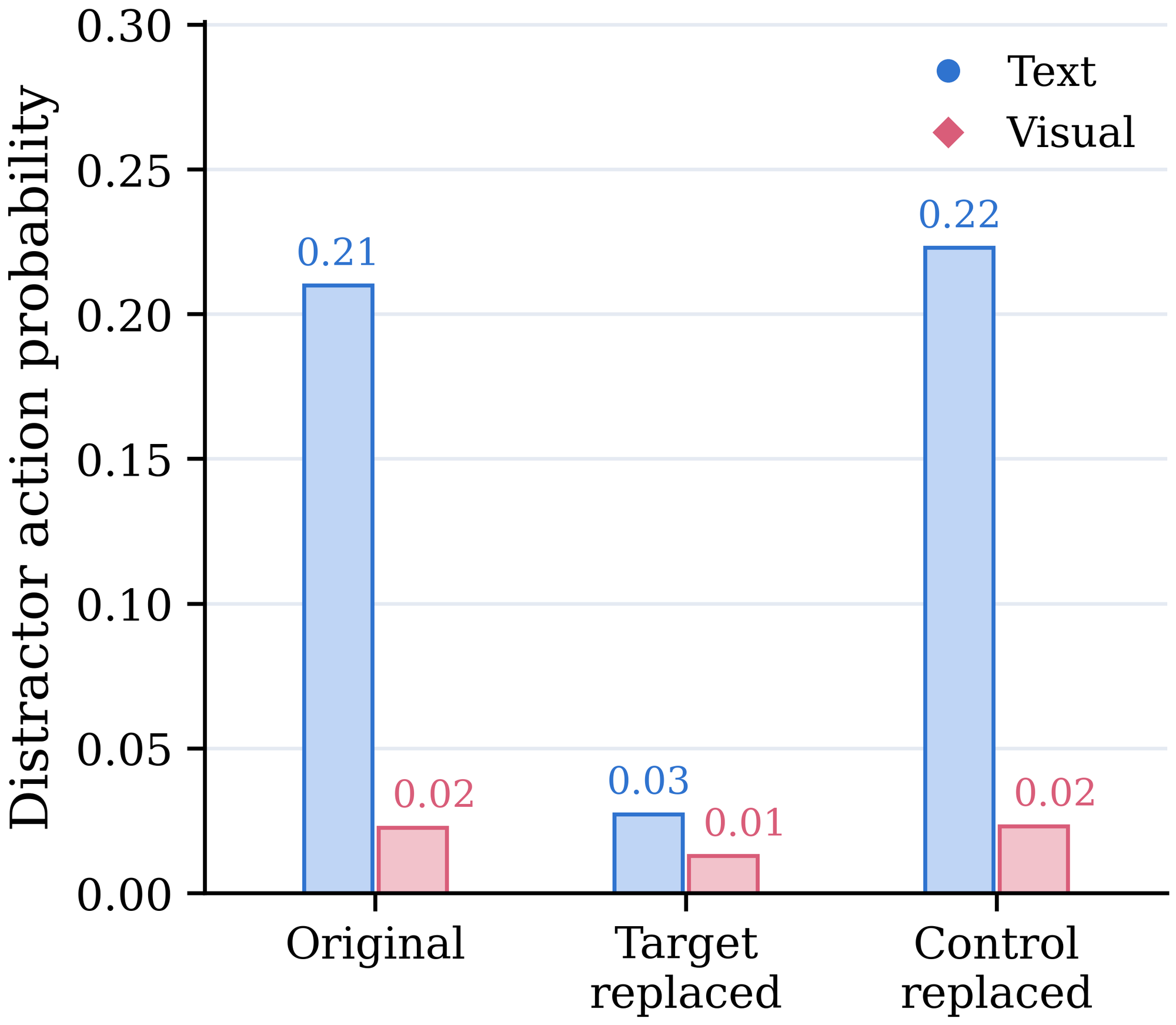}
        \end{minipage}
        \caption{Keyword-Replacement Controls.}
        \label{fig:procedural_exposure_bias_c_31}
    \end{subfigure}

    \caption{
        Procedural Exposure Bias in Gemma-4-31B-it Under Text and Visual Skill Representations.
        Figure~\ref{fig:procedural_exposure_bias_a_31}: Distractor Preference Across Exposure Levels.
        Figure~\ref{fig:procedural_exposure_bias_b_31}: Paired Preference Changes From $k=0$ to $k=8$.
        Figure~\ref{fig:procedural_exposure_bias_c_31}: Keyword-Replacement Controls.
    }
    \label{fig:procedural_exposure_bias_31}
\end{figure*}

Figures~\ref{fig:procedural_exposure_bias_27} and~\ref{fig:procedural_exposure_bias_31} report the same controlled probe for Qwen3.5-27B and Gemma-4-31B-it, respectively. Each figure includes exposure-dependent distractor preference, paired endpoint changes, and keyword-replacement controls.

\subsection{Details of Text and Visual Skill Encoding}
\label{app:representation_methodology}

% This section supplements Section~\ref{sec:methodology_representation} with the matched visual rendering procedure and examples of the structured skill cards used to present procedural content.

This section supplements Section~\ref{sec:methodology_representation} with details of the matched Text and Visual skill rendering procedure.

\subsubsection{Visual Reference Condition}
\label{app:peb_rendering}

Each exposure-controlled Text skill is paired with a Visual counterpart containing the same procedural wording. The interaction history, exposure level, and candidate actions are matched across the two conditions.

Text supplies the skill as language tokens. Visual renders it as a structured skill card and supplies the resulting image through the model's native visual encoding pathway, without adding auxiliary modules. Unless otherwise specified, Visual refers to this structured skill-card format. Because this comparison changes both the encoding pathway and the presentation of the skill, Appendix~\ref{app:rasterized_interference} additionally evaluates plain rasterized text without the skill-card structure.

\subsection{Details of the Experimental Setup}
\label{app:experimental_setup_details}

This section supplements Section~\ref{sec:experimental_setup} with benchmark descriptions, evaluation metrics, and the task-specific skill-ranking prompt. The complete-agent evaluation covers Qwen3.5-9B, Qwen3.5-27B, Gemma-4-31B-it, and Kimi-K2.5 under No Skill, Text Skill, and Visual Skill conditions.

\subsubsection{Datasets}
\label{app:benchmarks_metrics}
\label{app:datasets}

\paragraph{ALFWorld.}
ALFWorld~\citep{DBLP:conf/iclr/ShridharYCBTH21} aligns text-based household environments with embodied tasks. Agents follow natural-language goals by locating objects, interacting with appliances and receptacles, and changing object states. Its six task categories are Pick \& Place, Examine in Light, Clean \& Place, Heat \& Place, Cool \& Place, and Pick Two \& Place. Task success rate measures completion of the requested household goal. This task-level measure is distinct from partial goal-condition completion. Our task-category success rates and interaction steps are reported in Appendix~\ref{app:alfworld_task_results}.

\paragraph{ScienceWorld.}
ScienceWorld~\citep{wang2022scienceworld} is an interactive text environment for performing elementary science tasks. The benchmark covers a range of scientific topics, including changes in states of matter, electrical conductivity, and biological processes. Agents must carry out procedures in the environment rather than only answer factual questions. The original benchmark provides normalized scores based on required goals and optional subgoals, allowing partial progress to receive credit. Such scores differ from task success rates: the success rates reported in this paper concern completed tasks rather than average partial-progress scores. Task-level results are provided in Appendix~\ref{app:scienceworld_task_results}.

\paragraph{WebShop.}
WebShop~\citep{DBLP:conf/nips/Yao0YN22} is a simulated shopping environment in which agents search for products, inspect product information, select options, and make a purchase according to a natural-language instruction. Its reward evaluates the purchased product against requested attributes, options, price, and product type. The original benchmark distinguishes task score, defined as $100$ times mean reward, from success rate, defined as the proportion of episodes receiving reward $1$. The WebShop results in Table~\ref{tab:main_results} use success rate, consistent with the main evaluation.

\subsubsection{Evaluation Metrics}
\label{app:evaluation_metrics}

The main evaluation reports task success rates across ALFWorld, ScienceWorld, and WebShop. Success rate denotes the proportion of evaluated tasks completed successfully. For each model and skill condition, the macro-average in Table~\ref{tab:main_results} is the arithmetic mean of the three environment-level success rates, giving each environment equal weight. Task-level interaction steps are additionally reported for ALFWorld and ScienceWorld.

\subsubsection{Skill Ranking Prompt}
\label{app:skill_ranking_prompt}

As described in Section~\ref{sec:overall_evaluation}, Qwen3.8-Max ranks the shared skill library for each task instance using its type, name, and goal. Text and Visual use the same ranked skills and procedural content, supplied either as language tokens or as rendered skill cards. Table~\ref{tab:skill_ranking_prompt} presents the ALFWorld ranking prompt, whose placeholders contain task information and the available skill names and descriptions.

\begin{table}[t]
\centering
\small
\caption{
Prompt used to rank the shared skill library for each ALFWorld task instance.
The placeholders are instantiated with the task type, task name, task goal, and the names and descriptions of all available skills.
}
\setlength{\tabcolsep}{7pt}
\renewcommand{\arraystretch}{1.18}
\begin{tabularx}{0.96\linewidth}{|>{\raggedright\arraybackslash}X|}
\hline
\rowcolor{blue!10}
\centering Task Specific Skill Ranking Prompt
\tabularnewline
\hline

{\ttfamily
You are an expert in embodied AI agent task decomposition for the ALFWorld household environment.
}
\tabularnewline[5pt]

Task Instance
\tabularnewline[2pt]

{\ttfamily
Task type: \{task\_type\}\newline
Task name: \{task\_name\}\newline
Goal: \{query\}
}
\tabularnewline[5pt]

Available Skills (\{num\_skills\} total)
\tabularnewline[2pt]

{\ttfamily
\{skills\_text\}
}
\tabularnewline[5pt]

Instruction
\tabularnewline[2pt]

{\ttfamily
Rank ALL \{num\_skills\} skills by their relevance to completing this specific task instance: ``\{query\}''.\par
\vspace{3pt}
Assign each skill a relevance score from 1 (not relevant at all) to 10 (essential/critical for this task).\par
\vspace{3pt}
Consider which skills would be directly invoked during execution, which are supportive or auxiliary, and which are irrelevant.\par
\vspace{3pt}
Return a JSON array sorted from MOST relevant (highest score) to LEAST relevant (lowest score).
}
\tabularnewline[5pt]

Output Format
\tabularnewline[2pt]

{\ttfamily
[\newline
\{"skill\_name": "xxx", "relevance\_score": 10, "reason": "brief reason"\},\newline
\{"skill\_name": "xxx", "relevance\_score": 8, "reason": "brief reason"\},\newline
...\newline
]\par
\vspace{3pt}
Return ONLY the JSON array, with no markdown fences or additional text.
}
\tabularnewline

\hline
\end{tabularx}

\label{tab:skill_ranking_prompt}
\vspace{-10pt}
\end{table}

\subsection{Supplementary Complete-Agent Evaluation Results}
\label{app:complete_agent_evaluation}

This section supplements Section~\ref{sec:overall_evaluation} with task-level success rates and interaction-step results for ALFWorld and ScienceWorld.

\subsubsection{Task-Level Performance under the Default Skill Setting}
\label{app:task_level_results}

The following tables break down the complete-agent results by task category for ALFWorld and ScienceWorld, reporting success rates and interaction steps.

\paragraph{ALFWorld.}
\label{app:alfworld_task_results}
\begin{table*}[t]
\centering
\small
\setlength{\tabcolsep}{3.0pt}
\renewcommand{\arraystretch}{1.08}

\caption{Success rate (\%) across different ALFWorld task types under the default skill setting. \textbf{Overall} denotes the overall success rate. Higher is better ($\uparrow$). The best result for each task type within each backbone is highlighted in \textbf{bold}.}

\begin{tabularx}{0.94\textwidth}{@{}>{\centering\arraybackslash}p{1.55cm}*{7}{>{\centering\arraybackslash}X}@{}}
\toprule
\textbf{Setting}
& \mbox{\textbf{Overall}~$\uparrow$}
& \mbox{\textbf{Pick}~$\uparrow$}
& \mbox{\textbf{Clean}~$\uparrow$}
& \mbox{\textbf{Heat}~$\uparrow$}
& \mbox{\textbf{Cool}~$\uparrow$}
& \mbox{\textbf{Pick2}~$\uparrow$}
& \mbox{\textbf{Examine}~$\uparrow$} \\
\midrule

\multicolumn{8}{c}{\textbf{Qwen3.5-9B}} \\
\hdashline
No Skill
& 15.67
& 33.33
& 6.45
& 4.35
& 4.76
& 11.76
& 38.89 \\

Text Skill
& 66.42
& 66.67
& \textbf{65.62}
& \textbf{57.14}
& \textbf{77.27}
& 56.25
& \textbf{73.68} \\

\rowcolor{cyan!3} Visual Skill
& \textbf{70.15}
& \textbf{91.67}
& 64.29
& 50.00
& \textbf{77.27}
& \textbf{88.24}
& 52.63 \\
\midrule

\multicolumn{8}{c}{\textbf{Qwen3.5-27B}} \\
\hdashline
No Skill
& 58.96
& 87.50
& 80.65
& 4.35
& 14.29
& 82.35
& \textbf{83.33} \\

Text Skill
& 78.36
& \textbf{91.67}
& 80.65
& \textbf{65.22}
& \textbf{95.24}
& 70.59
& 61.11 \\

\rowcolor{cyan!3} Visual Skill
& \textbf{80.60}
& \textbf{91.67}
& \textbf{93.55}
& 56.52
& \textbf{95.24}
& \textbf{94.12}
& 44.44 \\
\midrule

\multicolumn{8}{c}{\textbf{Gemma4-31B}} \\
\hdashline
No Skill
& 62.69
& 95.83
& 70.97
& 0.00
& 76.19
& 64.71
& 66.67 \\

Text Skill
& \textbf{90.30}
& \textbf{100.00}
& 93.55
& \textbf{78.26}
& \textbf{95.24}
& \textbf{94.12}
& 77.78 \\

\rowcolor{cyan!3} Visual Skill
& \textbf{90.30}
& \textbf{100.00}
& \textbf{96.77}
& 69.57
& 90.48
& 88.24
& \textbf{94.44} \\
\midrule

\multicolumn{8}{c}{\textbf{Kimi-K2.5}} \\
\hdashline
No Skill
& 78.36
& 95.83
& 87.10
& 30.43
& 76.19
& 82.35
& \textbf{100.00} \\

Text Skill
& 90.30
& 95.83
& 90.32
& \textbf{65.22}
& \textbf{95.24}
& \textbf{100.00}
& \textbf{100.00} \\

\rowcolor{cyan!3} Visual Skill
& \textbf{91.79}
& \textbf{100.00}
& \textbf{93.55}
& \textbf{65.22}
& \textbf{95.24}
& \textbf{100.00}
& \textbf{100.00} \\

\bottomrule
\end{tabularx}

\label{tab:alfworld_success_by_task}
\vspace{-8pt}
\end{table*}

\begin{table*}[t]
\centering
\small
\setlength{\tabcolsep}{3.0pt}
\renewcommand{\arraystretch}{1.08}
\caption{Average number of interaction steps across different ALFWorld task types under the default skill setting. \textbf{Overall} denotes the average number of interaction steps across all tasks. Lower is better ($\downarrow$). The best result for each task type within each backbone is highlighted in \textbf{bold}.}
\begin{tabularx}{0.94\textwidth}{@{}>{\centering\arraybackslash}p{1.55cm}*{7}{>{\centering\arraybackslash}X}@{}}
\toprule
\textbf{Setting} & \mbox{\textbf{Overall}~$\downarrow$} & \mbox{\textbf{Pick}~$\downarrow$} & \mbox{\textbf{Clean}~$\downarrow$} & \mbox{\textbf{Heat}~$\downarrow$} & \mbox{\textbf{Cool}~$\downarrow$} & \mbox{\textbf{Pick2}~$\downarrow$} & \mbox{\textbf{Examine}~$\downarrow$} \\
\midrule
\multicolumn{8}{c}{\textbf{Qwen3.5-9B}} \\
\hdashline
No Skill & 27.59 & 25.04 & 29.45 & 29.78 & 29.62 & 27.65 & 22.56 \\
Text Skill & 18.28 & 17.88 & \textbf{18.69} & \textbf{21.05} & 18.18 & 20.81 & \textbf{13.00} \\
\rowcolor{cyan!3} Visual Skill & \textbf{17.75} & \textbf{9.96} & 19.25 & 21.83 & \textbf{17.91} & \textbf{16.94} & 20.74 \\
\midrule
\multicolumn{8}{c}{\textbf{Qwen3.5-27B}} \\
\hdashline
No Skill & 19.01 & 11.21 & 15.32 & 29.17 & 27.48 & 17.47 & \textbf{14.39} \\
Text Skill & 14.87 & \textbf{10.33} & 15.87 & \textbf{16.48} & 13.67 & 17.12 & 16.39 \\
\rowcolor{cyan!3} Visual Skill & \textbf{14.66} & 11.67 & \textbf{13.16} & 19.87 & \textbf{11.76} & \textbf{13.18} & 19.39 \\
\midrule
\multicolumn{8}{c}{\textbf{Gemma4-31B}} \\
\hdashline
No Skill & 18.66 & 9.38 & 17.55 & 30.00 & 15.81 & 22.76 & 17.89 \\
Text Skill & 12.01 & 9.79 & \textbf{11.84} & 17.13 & 10.19 & 13.71 & \textbf{9.28} \\
\rowcolor{cyan!3} Visual Skill & \textbf{11.78} & \textbf{8.04} & 12.81 & \textbf{13.96} & \textbf{9.14} & \textbf{12.71} & 14.39 \\
\midrule
\multicolumn{8}{c}{\textbf{Kimi-K2.5}} \\
\hdashline
No Skill & 16.76 & 9.75 & 16.35 & 26.26 & 17.29 & 20.24 & 10.78 \\
Text Skill & 12.75 & \textbf{8.50} & \textbf{13.13} & 22.26 & 10.67 & 14.29 & \textbf{6.61} \\
\rowcolor{cyan!3} Visual Skill & \textbf{12.41} & \textbf{8.50} & 14.71 & \textbf{17.83} & \textbf{10.29} & \textbf{12.65} & 9.00 \\
\bottomrule
\end{tabularx}
\label{tab:alfworld_steps_by_task}
\vspace{-8pt}
\end{table*}

Tables~\ref{tab:alfworld_success_by_task} and~\ref{tab:alfworld_steps_by_task} report success rates and interaction steps across the six ALFWorld task categories. The task-level breakdown shows that the performance difference between Text and Visual is not uniform across categories. Visual's relative advantage varies with the task category, indicating that the aggregate result reflects different contributions from individual task types. The interaction-step results provide a complementary view of the number of steps used under the two representations.

\paragraph{ScienceWorld.}
\label{app:scienceworld_task_results}
\begin{table*}[t]
\centering
\small
\setlength{\tabcolsep}{3.0pt}
\renewcommand{\arraystretch}{1.08}
\caption{Success rate (\%) across different ScienceWorld task types under the default skill setting. \textbf{Overall} denotes the overall success rate. Higher is better ($\uparrow$). The best result for each task type within each backbone is highlighted in \textbf{bold}.}
\begin{tabularx}{0.98\textwidth}{@{}>{\centering\arraybackslash}p{1.55cm}*{9}{>{\centering\arraybackslash}X}@{}}
\toprule
\textbf{Setting} & \mbox{\textbf{Overall}~$\uparrow$} & \mbox{\textbf{State}~$\uparrow$} & \mbox{\textbf{Elec.}~$\uparrow$} & \mbox{\textbf{Class.}~$\uparrow$} & \mbox{\textbf{Growth}~$\uparrow$} & \mbox{\textbf{Chem.}~$\uparrow$} & \mbox{\textbf{Life}~$\uparrow$} & \mbox{\textbf{Stages}~$\uparrow$} & \mbox{\textbf{Meas.}~$\uparrow$} \\
\midrule
\multicolumn{10}{c}{\textbf{Qwen3.5-9B}} \\
\hdashline
No Skill & 18.96 & 0.00 & 36.67 & 7.50 & 5.00 & 11.54 & \textbf{70.00} & 0.00 & 5.00 \\
Text Skill & 42.18 & \textbf{8.33} & 20.00 & \textbf{95.00} & \textbf{40.00} & 19.23 & 56.67 & 11.11 & \textbf{55.00} \\
\rowcolor{cyan!3} Visual Skill & \textbf{45.02} & 5.56 & \textbf{46.67} & 92.50 & 25.00 & \textbf{30.77} & 60.00 & \textbf{22.22} & 45.00 \\
\midrule
\multicolumn{10}{c}{\textbf{Qwen3.5-27B}} \\
\hdashline
No Skill & 45.02 & 19.44 & 30.00 & 70.00 & 35.00 & 19.23 & \textbf{73.33} & \textbf{22.22} & \textbf{75.00} \\
Text Skill & 57.82 & \textbf{33.33} & 50.00 & \textbf{100.00} & \textbf{45.00} & 42.31 & 70.00 & \textbf{22.22} & 60.00 \\
\rowcolor{cyan!3} Visual Skill & \textbf{59.72} & 27.78 & \textbf{56.67} & \textbf{100.00} & \textbf{45.00} & \textbf{53.85} & \textbf{73.33} & \textbf{22.22} & 60.00 \\
\midrule
\multicolumn{10}{c}{\textbf{Gemma-4-31B}} \\
\hdashline
No Skill & 40.76 & \textbf{41.67} & 46.67 & 50.00 & 0.00 & 7.69 & 63.33 & 22.22 & \textbf{70.00} \\
Text Skill & 58.29 & 36.11 & 33.33 & \textbf{100.00} & \textbf{40.00} & \textbf{57.69} & \textbf{73.33} & \textbf{44.44} & 55.00 \\
\rowcolor{cyan!3} Visual Skill & \textbf{59.72} & \textbf{41.67} & \textbf{56.67} & 92.50 & \textbf{40.00} & 50.00 & \textbf{73.33} & \textbf{44.44} & 50.00 \\
\midrule
\multicolumn{10}{c}{\textbf{Kimi-K2.5}} \\
\hdashline
No Skill & 48.82 & \textbf{36.11} & 30.00 & 77.50 & 25.00 & 30.77 & 73.33 & 11.11 & \textbf{70.00} \\
Text Skill & 50.71 & 22.22 & \textbf{40.00} & \textbf{95.00} & 35.00 & \textbf{50.00} & 60.00 & 0.00 & 55.00 \\
\rowcolor{cyan!3} Visual Skill & \textbf{52.13} & 27.78 & 30.00 & 87.50 & \textbf{45.00} & 34.62 & \textbf{76.67} & \textbf{44.44} & 55.00 \\
\bottomrule
\end{tabularx}
\label{tab:scienceworld_success_by_task}
\vspace{-8pt}
\end{table*}
\begin{table*}[t]
\centering
\small
\setlength{\tabcolsep}{3.0pt}
\renewcommand{\arraystretch}{1.08}
\caption{Average number of interaction steps across different ScienceWorld task types under the default skill setting. \textbf{Overall} denotes the average number of interaction steps across all tasks. Lower is better ($\downarrow$). The best result for each task type within each backbone is highlighted in \textbf{bold}.}
\begin{tabularx}{0.98\textwidth}{@{}>{\centering\arraybackslash}p{1.65cm}*{9}{>{\centering\arraybackslash}X}@{}}
\toprule
\textbf{Setting} & \mbox{\textbf{Overall}~$\downarrow$} & \mbox{\textbf{State}~$\downarrow$} & \mbox{\textbf{Elec.}~$\downarrow$} & \mbox{\textbf{Class.}~$\downarrow$} & \mbox{\textbf{Growth}~$\downarrow$} & \mbox{\textbf{Chem.}~$\downarrow$} & \mbox{\textbf{Life}~$\downarrow$} & \mbox{\textbf{Stages}~$\downarrow$} & \mbox{\textbf{Meas.}~$\downarrow$} \\
\midrule
\multicolumn{10}{c}{\textbf{Qwen3.5-9B}} \\
\hdashline
No Skill & 17.04 & 21.62 & \textbf{22.57} & 12.25 & 22.80 & 25.36 & 4.97 & 21.02 & \textbf{9.80} \\
Text Skill & \textbf{15.36} & \textbf{20.22} & 22.67 & \textbf{7.53} & \textbf{19.00} & 24.77 & 4.37 & 7.67 & 15.40 \\
\rowcolor{cyan!3} Visual Skill & 15.95 & 20.23 & 24.60 & 9.38 & 22.45 & \textbf{23.02} & \textbf{3.67} & \textbf{7.56} & 14.95 \\
\midrule
\multicolumn{10}{c}{\textbf{Qwen3.5-27B}} \\
\hdashline
No Skill & 15.83 & \textbf{13.78} & 26.27 & 9.85 & 20.85 & \textbf{20.62} & 5.97 & 18.67 & 18.10 \\
Text Skill & 14.28 & 17.30 & \textbf{20.57} & \textbf{7.00} & 19.25 & 26.05 & \textbf{3.30} & 9.44 & 12.35 \\
\rowcolor{cyan!3} Visual Skill & \textbf{13.80} & 17.62 & 20.80 & 7.35 & \textbf{18.90} & 23.28 & \textbf{3.30} & \textbf{5.67} & \textbf{11.25} \\
\midrule
\multicolumn{10}{c}{\textbf{Gemma-4-31B}} \\
\hdashline
No Skill & 16.23 & \textbf{15.15} & 25.93 & 10.90 & 24.70 & \textbf{19.01} & 6.53 & 18.89 & \textbf{15.50} \\
Text Skill & 14.99 & 18.73 & 23.93 & \textbf{7.55} & 20.45 & 22.31 & \textbf{3.33} & 6.56 & 16.05 \\
\rowcolor{cyan!3} Visual Skill & \textbf{14.60} & 17.62 & \textbf{23.53} & 8.07 & \textbf{19.35} & 20.96 & \textbf{3.33} & \textbf{6.47} & 16.30 \\
\midrule
\multicolumn{10}{c}{\textbf{Kimi-K2.5}} \\
\hdashline
No Skill & 16.54 & \textbf{17.47} & 27.47 & \textbf{8.45} & 24.30 & \textbf{20.77} & 5.50 & 18.33 & 17.15 \\
Text Skill & \textbf{15.95} & 21.61 & \textbf{19.60} & 9.07 & 23.20 & 23.27 & \textbf{3.37} & 11.44 & 18.15 \\
\rowcolor{cyan!3} Visual Skill & 16.06 & 20.56 & 20.13 & 9.75 & \textbf{23.05} & 26.27 & 3.80 & \textbf{10.11} & \textbf{15.30} \\
\bottomrule
\end{tabularx}
\label{tab:scienceworld_steps_by_task}
\vspace{-8pt}
\end{table*}

Tables~\ref{tab:scienceworld_success_by_task} and~\ref{tab:scienceworld_steps_by_task} report the corresponding task-level results for ScienceWorld. The subtasks contributing to the aggregate Text--Visual performance difference vary across model backbones. Thus, the task-level pattern is model-dependent, even when the aggregate comparison favors the same representation. The success-rate and interaction-step breakdowns show how the representation differences vary across individual scientific tasks.

\subsection{Supplementary Representation-Dependent Exposure Analyses}
\label{app:representation_experiments}
\label{app:visual_encoding_mechanism}

This section supplements Section~\ref{sec:representation_analysis} with the rasterized representation comparison, repeated-signal readout, executable action selection, paired rollouts, layer-wise readout, and component interventions. Unless otherwise specified, these analyses use the 165 decision states and exposure-controlled variants described in Appendix~\ref{app:peb_details}. The executable-action experiment uses a separate set of states, described below.

\subsubsection{Rasterized Representation Intervention}
\label{app:rasterized_interference}

This experiment examines whether the attenuation observed under Visual input also occurs without the additional structure of the skill-card layout.

\paragraph{Skill Representations.}
We compare three representations of the same procedural content. \textit{Text} supplies the skill as language tokens. \textit{Rasterized} renders the same wording and ordering as a plain monospaced image without additional structural formatting. \textit{Structural Visual} uses a separately implemented structured skill-card format based on the main-experiment design. Comparing Text with Rasterized assesses the representation change without additional skill-card structure. Comparing Rasterized with Structural Visual assesses the additional effect of that layout.

\paragraph{Exposure Sensitivity.}
For this analysis, we use the summed action-token log-probability $\widetilde{S}$, rather than the length-normalized score $S$:
\begin{equation}
\widetilde{S}_i(a;k,r)
=
\sum_{t=1}^{|a|}
\log p_{\theta}
\left(a_t \mid x_i^{(k,r)},a_{<t}\right).
\label{eq:ri_action_score}
\end{equation}
For each state and representation, exposure sensitivity is estimated using the linear fit
\begin{equation}
\widetilde{S}_i(a_i^{D};k,r)
=
\alpha_{i,r}
+
\beta_{i,r}k
+
\epsilon_{i,k,r},
\label{eq:ri_exposure_sensitivity}
\end{equation}
where $\beta_{i,r}$ summarizes the change in distractor score with exposure over $k\in\{0,1,2,4,8\}$.

\begin{figure*}[t]
    \centering

    \begin{subfigure}[t]{0.48\textwidth}
        \centering
        \includegraphics[width=\linewidth]{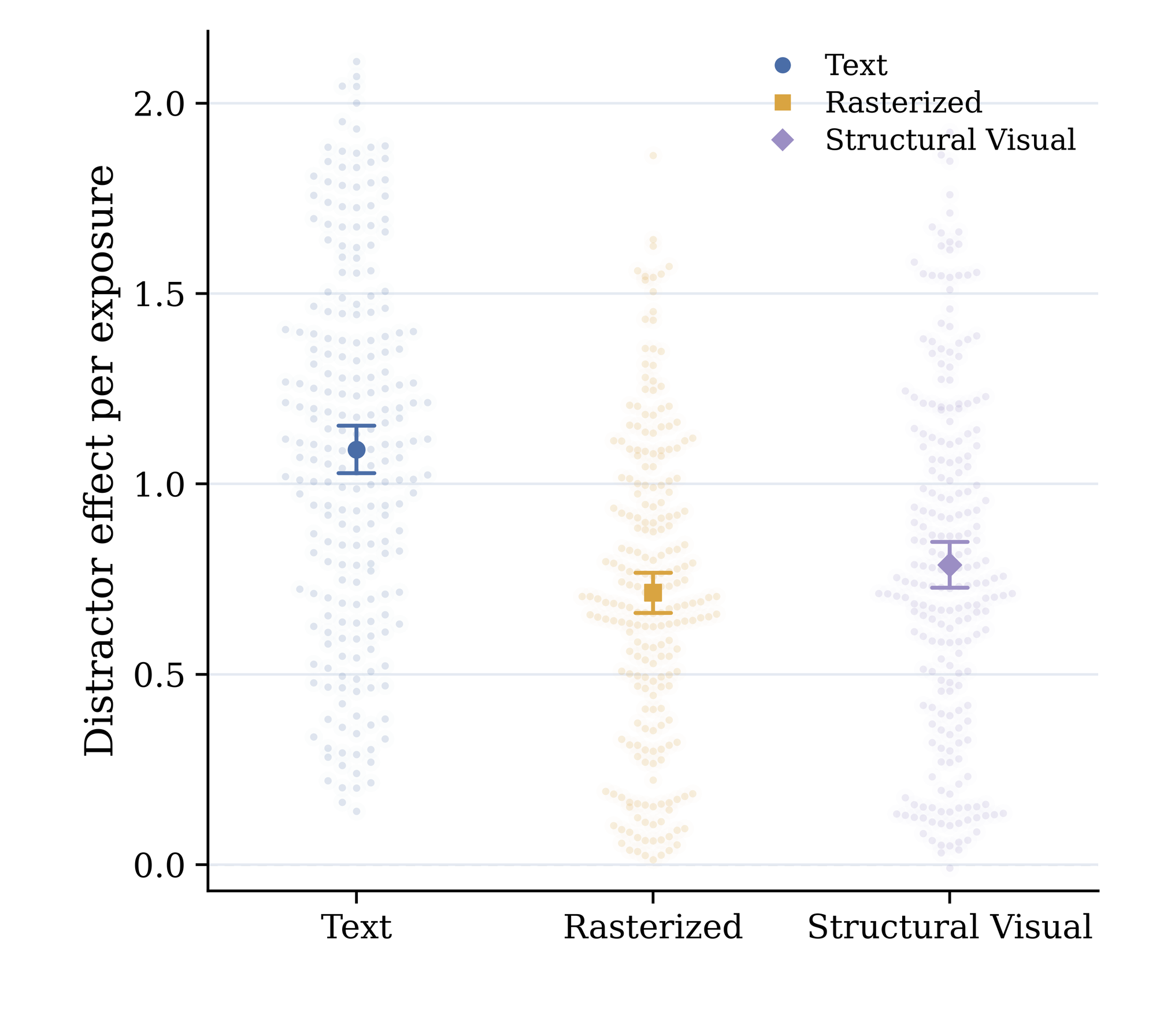}
        \caption{Qwen3.5-27B.}
        \label{fig:ri_27b}
    \end{subfigure}
    \hfill
    \begin{subfigure}[t]{0.48\textwidth}
        \centering
        \includegraphics[width=\linewidth]{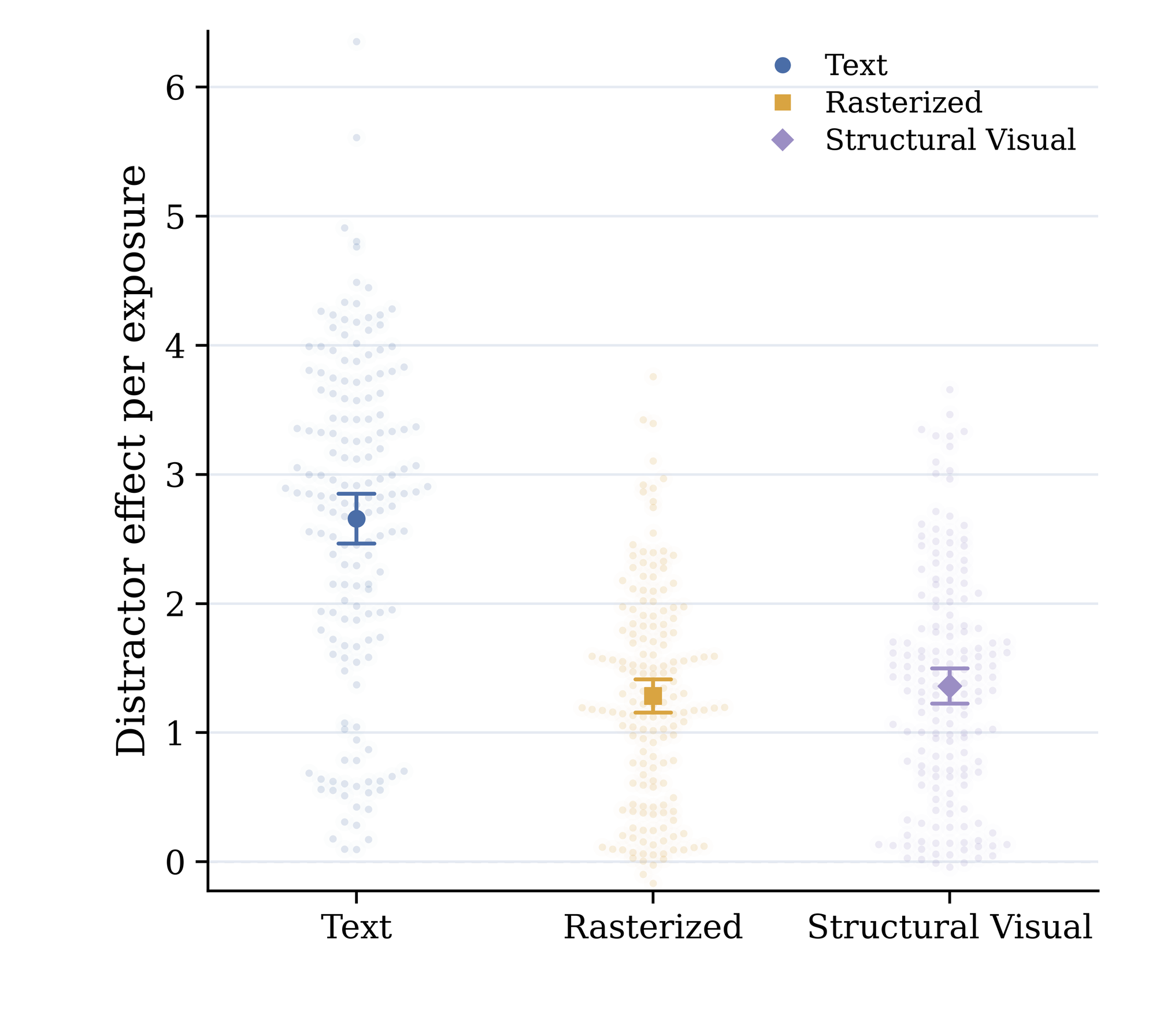}
        \caption{Gemma-4-31B-it.}
        \label{fig:ri_31b}
    \end{subfigure}

    \caption{
        Additional Results for the Rasterized Representation Intervention.
        Each panel reports state-level exposure sensitivity under Text, Rasterized, and Structural Visual representations. Central markers and whiskers denote the mean and 95\% confidence interval.
    }
    \label{fig:ri_additional_models}
\end{figure*}

For Qwen3.5-9B, mean exposure sensitivity decreases from $0.741$ under Text to $0.467$ under Rasterized input, and increases to $0.552$ under Structural Visual input. Gemma-4-31B-it shows the same ordering, with corresponding values of $2.657$, $1.283$, and $1.361$. Substantial attenuation is thus already observed under plain image rendering, without the structured skill-card layout. Figure~\ref{fig:ri_additional_models} reports the results for Qwen3.5-27B and Gemma-4-31B-it.

\subsubsection{Repeated Signal Readout}
\label{app:signal_readout}

This probe examines how repeated procedural information influences target-word prediction under visual encoding by replacing action prediction with a direct target-word readout.

\paragraph{Setup and Scoring.}
We use variants in which the target object word occurs $k\in\{0,4,8\}$ times and prompt the model to output the most frequent object in the skill. The target word is scored at a fixed answer position. Let $x_{i,\mathrm{read}}^{(k,r)}$ denote the input containing the skill and the readout prompt, and let the target word $w$ tokenize into $(u_1,\ldots,u_m)$. Its sequence log-probability is
\begin{equation}
R_i(w;k,r)
=
\sum_{j=1}^{m}
\log p_{\theta}
\left(u_j \mid x_{i,\mathrm{read}}^{(k,r)},u_{<j}\right).
\label{eq:symbol_readout_score}
\end{equation}
The corresponding target-token-sequence probability is
\begin{equation}
P_i^{\mathrm{read}}(w;k,r)
=
\exp\!\left(R_i(w;k,r)\right).
\label{eq:symbol_readout_probability}
\end{equation}
This metric measures support for the target word at the designated answer position; it is not an exact-match accuracy measure for the most-frequent-object task.

\paragraph{Readout Results and Control.}
Figure~\ref{fig:target_word_readout} reports results for Qwen3.5-9B and Gemma-4-31B-it. Target-word probability increases with exposure under both Text and Visual, indicating that repetition continues to influence target-word prediction after visual encoding. This does not establish equal accessibility across representations or rule out partial degradation of the signal.

An unrelated frequency-matched control word is scored using the same procedure at $k\in\{4,8\}$. The control probabilities remain approximately unchanged in the reported comparison, providing a check against a general increase in output confidence.

\subsubsection{Executable Action Selection}
\label{app:action_selection_details}

This experiment tests whether exposure-dependent preference shifts extend to freely generated actions when the competing action is admissible in the current environment state.

\paragraph{Decision States and Competing Actions.}
We construct a separate set of 185 decision states from 64 successful ALFWorld trajectories. For each state, $a_i^{C}$ denotes the reference-trajectory action, and $a_i^{\mathrm{comp}}$ denotes a task-relevant competing action selected from the current admissible action set. The competing action is executable but differs from the reference action.

\paragraph{Exposure-Controlled Generation.}
The context contains five ranked skills. We manipulate the keyword associated with the competing action over $k\in\{0,1,2,4,8\}$ using fixed insertion slots, while preserving content outside those slots and matching word counts across variants. The model freely generates the next action using greedy decoding under Text or Visual input.

For representation $r$, the competing-action selection rate is
\begin{equation}
R_r(k)
=
\frac{1}{N}
\sum_{i=1}^{N}
\mathbb{I}\!\left[
\hat{a}_{i,r}^{(k)}=a_i^{\mathrm{comp}}
\right],
\label{eq:behavior_selection_rate}
\end{equation}
where $N=185$ and $\hat{a}_{i,r}^{(k)}$ is the generated action. Figure~\ref{fig:behavioral_realization} reports the selection-rate results for Qwen3.5-9B and Gemma-4-31B-it.

\paragraph{Aggregation and Statistical Analysis.}
We summarize the aggregate relationship between exposure and selection rate using
\begin{equation}
R_r(k)
=
\alpha_r+\beta_r k+\epsilon_{r,k}.
\label{eq:behavior_group_slope}
\end{equation}
The reported aggregate fits yield a Text slope of $0.0091$ ($R^2=0.95$) and a Visual slope of $0.0026$ ($R^2=0.48$). The paired slope difference is $0.0065$, with a bootstrap 95\% confidence interval of $[0.0011,0.0120]$. These fits summarize the aggregate selection-rate trend and do not imply monotonic switching at every decision state.

\subsubsection{Paired Rollout Evaluation}
\label{app:paired_rollouts}

\begin{table*}[t]
\caption{
Paired Rollout Results for Qwen3.5-9B on ALFWorld under the Text Skill Representation.
The Text condition is evaluated at $k\in\{0,4,8\}$.
Success rates and mean continuation steps are computed over all 185 decision states for each branch and exposure level.
$\Delta L$ denotes the competing-minus-correct continuation-step difference among pairs for which both branches successfully complete the task.
Confidence intervals use 4,000 game-clustered bootstrap resamples, and the final column reports the exact McNemar test for paired rollout success.
The Pooled row aggregates the both-success pairs across the three exposure levels.
}
\centering
\small
\setlength{\tabcolsep}{4.8pt}
\renewcommand{\arraystretch}{1.12}
\begin{tabular}{c cc cc c c c}
\toprule
& \multicolumn{2}{c}{Success Rate (\%)} & \multicolumn{2}{c}{Mean Steps} & \multicolumn{2}{c}{Both-Success Analysis} & \\
\cmidrule(lr){2-3}\cmidrule(lr){4-5}\cmidrule(lr){6-7}
$k$ & Correct & Competing & Correct & Competing & Pairs & $\Delta L$ [95\% CI] & McNemar $p$ \\
\midrule
0 & 30.27 & 28.11 & 7.60 & 9.09 & 25 & $+2.52\ [1.48,\,3.97]$ & 0.694 \\
4 & 40.54 & 30.27 & 8.09 & 8.95 & 33 & $+2.97\ [1.69,\,4.75]$ & 0.025 \\
8 & 34.05 & 26.49 & 8.06 & 8.44 & 28 & $+1.36\ [0.48,\,2.14]$ & 0.081 \\
\midrule
Pooled & N/A & N/A & N/A & N/A & 86 & $+2.31\ [1.61,\,3.21]$ & N/A \\
\bottomrule
\end{tabular}
\label{tab:paired_rollout_results}
\end{table*}

\begin{table*}[t]
\caption{
Paired Rollout Results for Qwen3.5-9B on ALFWorld under the Visual Skill Representation.
The Visual condition is evaluated at $k\in\{0,4,8\}$ using the structured skill-card representation.
Success rates and mean continuation steps are computed over all 185 decision states for each branch and exposure level.
$\Delta L$ denotes the competing-minus-correct continuation-step difference among pairs for which both branches successfully complete the task.
Confidence intervals use 4,000 game-clustered bootstrap resamples, and the final column reports the exact McNemar test for paired rollout success.
The Pooled row aggregates the both-success pairs across the three exposure levels.
}
\centering
\small
\setlength{\tabcolsep}{4.8pt}
\renewcommand{\arraystretch}{1.12}
\begin{tabular}{c cc cc c c c}
\toprule
& \multicolumn{2}{c}{Success Rate (\%)} & \multicolumn{2}{c}{Mean Steps} & \multicolumn{2}{c}{Both-Success Analysis} & \\
\cmidrule(lr){2-3}\cmidrule(lr){4-5}\cmidrule(lr){6-7}
$k$ & Correct & Competing & Correct & Competing & Pairs & $\Delta L$ [95\% CI] & McNemar $p$ \\
\midrule
0 & 25.41 & 31.35 & 16.32 & 17.06 & 24 & $+2.17\ [0.13,\,4.27]$ & 0.185 \\
4 & 41.08 & 31.35 & 15.08 & 16.97 & 36 & $+3.44\ [1.75,\,5.41]$ & 0.030 \\
8 & 39.46 & 31.89 & 16.31 & 16.39 & 38 & $+1.63\ [0.10,\,3.27]$ & 0.081 \\
\midrule
Pooled & N/A & N/A & N/A & N/A & 98 & $+2.43\ [1.26,\,3.90]$ & N/A \\
\bottomrule
\end{tabular}
\label{tab:paired_rollout_results_visual}
\end{table*}

This evaluation examines the downstream consequences of executing the competing action rather than the reference action from a matched initial state.

\paragraph{Paired Continuations.}
For each evaluated state, we create two environment branches. The \emph{reference} branch first executes $a_i^{C}$, while the \emph{competing} branch first executes $a_i^{\mathrm{comp}}$. Both branches then continue with unrestricted greedy action generation under the same skill representation and exposure condition within the pair.

Rollouts terminate upon task success, an unparsable action, or exhaustion of the state-dependent budget
\begin{equation}
B_i
=
\min\!\left(40,\;B_i^{\mathrm{expert}}+20\right),
\label{eq:rollout_budget}
\end{equation}
where $B_i^{\mathrm{expert}}$ is the number of remaining steps in the reference trajectory from decision step $i$ onward.

\paragraph{Conditional Step Difference.}
For pairs in which both branches succeed, we compute
\begin{equation}
\Delta L_{i,r}(k)
=
L_{i,r,k}^{\mathrm{comp}}
-
L_{i,r,k}^{C},
\label{eq:paired_rollout_cost}
\end{equation}
where $L_{i,r,k}^{\mathrm{comp}}$ and $L_{i,r,k}^{C}$ are the recorded rollout step counts for the competing and reference branches. A positive value indicates additional steps in the competing branch. This metric is conditional on both branches succeeding and does not capture differences in failure rates.

Tables~\ref{tab:paired_rollout_results} and~\ref{tab:paired_rollout_results_visual} report the paired outcomes under Text and Visual, respectively. The paired-rollout comparison assesses the consequences of the initial action choice; it does not by itself estimate the effect of exposure on complete-agent performance.

\subsubsection{Layer-Wise Readout of Exposure-Dependent Action Evidence}
\label{app:layerwise_readout}

We use intermediate-layer readouts to examine how exposure-related action evidence varies across model depth.

\paragraph{Experimental Setup.}
We perform teacher-forced forward passes for the reference and distractor actions. Let $q_{i,k,r,t}$ denote the input position immediately preceding action token $a_t$ in the corresponding sequence. At language-model layer $l$, the hidden state at this prediction position is passed through the model's final normalization and language-model head. The layer-wise action score is
\begin{equation}
S_{i,l}(a;k,r)
=
\frac{1}{|a|}
\sum_{t=1}^{|a|}
\log
\left[
\mathrm{softmax}
\left(
W_{\mathrm{LM}}\mathrm{Norm}
\left(h^{(l)}_{i,k,r,q_{i,k,r,t}}\right)
\right)
\right]_{a_t}.
\label{eq:layerwise_action_score}
\end{equation}
Here, $W_{\mathrm{LM}}$ denotes the language-model output projection, and $\mathrm{Norm}$ denotes the final normalization. These scores are intermediate readouts under teacher forcing rather than probabilities from independent generation at each layer.

\begin{figure*}[t]
    \centering

    \begin{subfigure}[t]{0.235\textwidth}
        \centering
        \includegraphics[width=\linewidth]{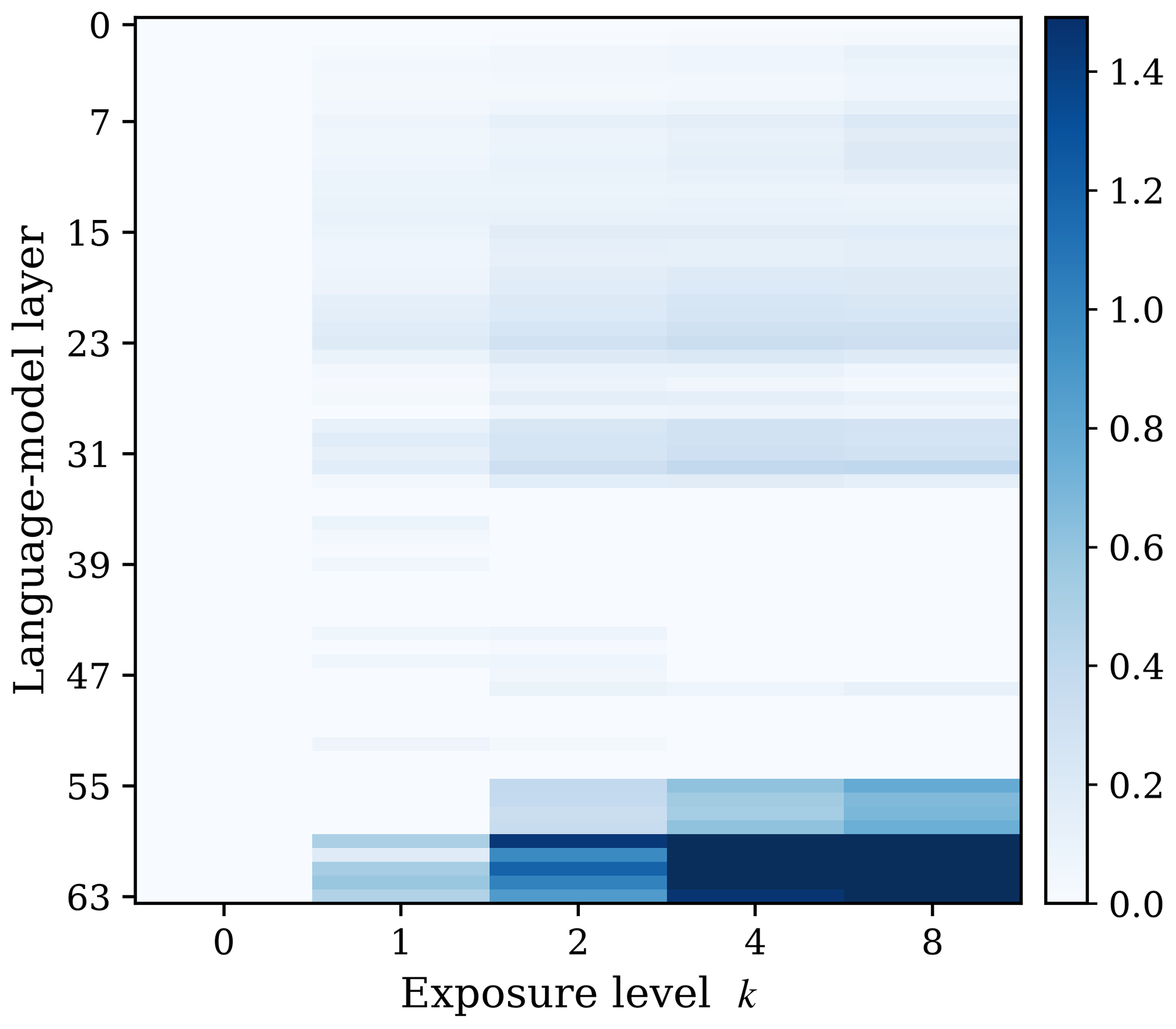}
        \caption{Qwen3.5-27B, Text.}
        \label{fig:layerwise_27b_text}
    \end{subfigure}
    \hfill
    \begin{subfigure}[t]{0.235\textwidth}
        \centering
        \includegraphics[width=\linewidth]{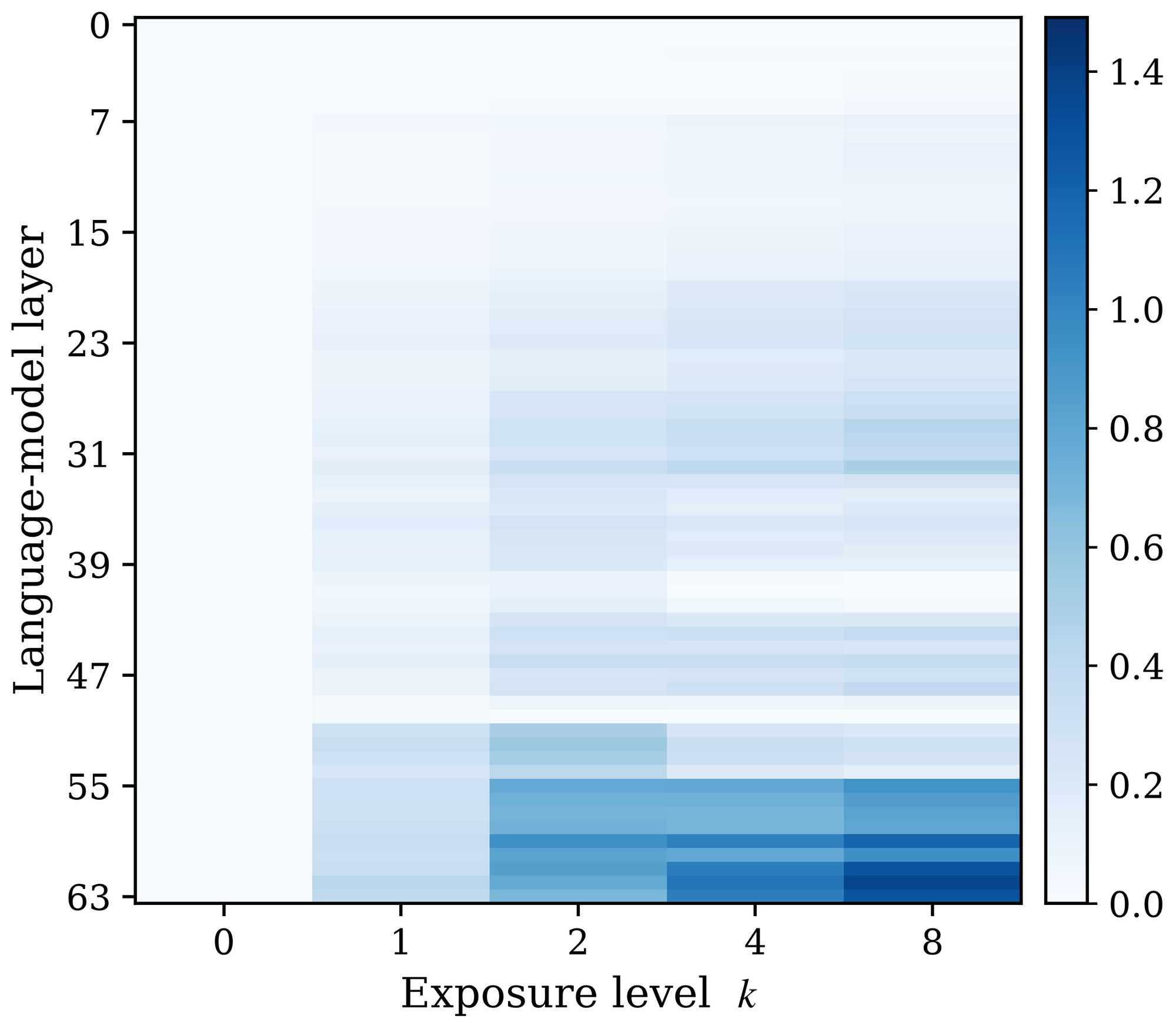}
        \caption{Qwen3.5-27B, Visual.}
        \label{fig:layerwise_27b_visual}
    \end{subfigure}
    \hfill
    \begin{subfigure}[t]{0.235\textwidth}
        \centering
        \includegraphics[width=\linewidth]{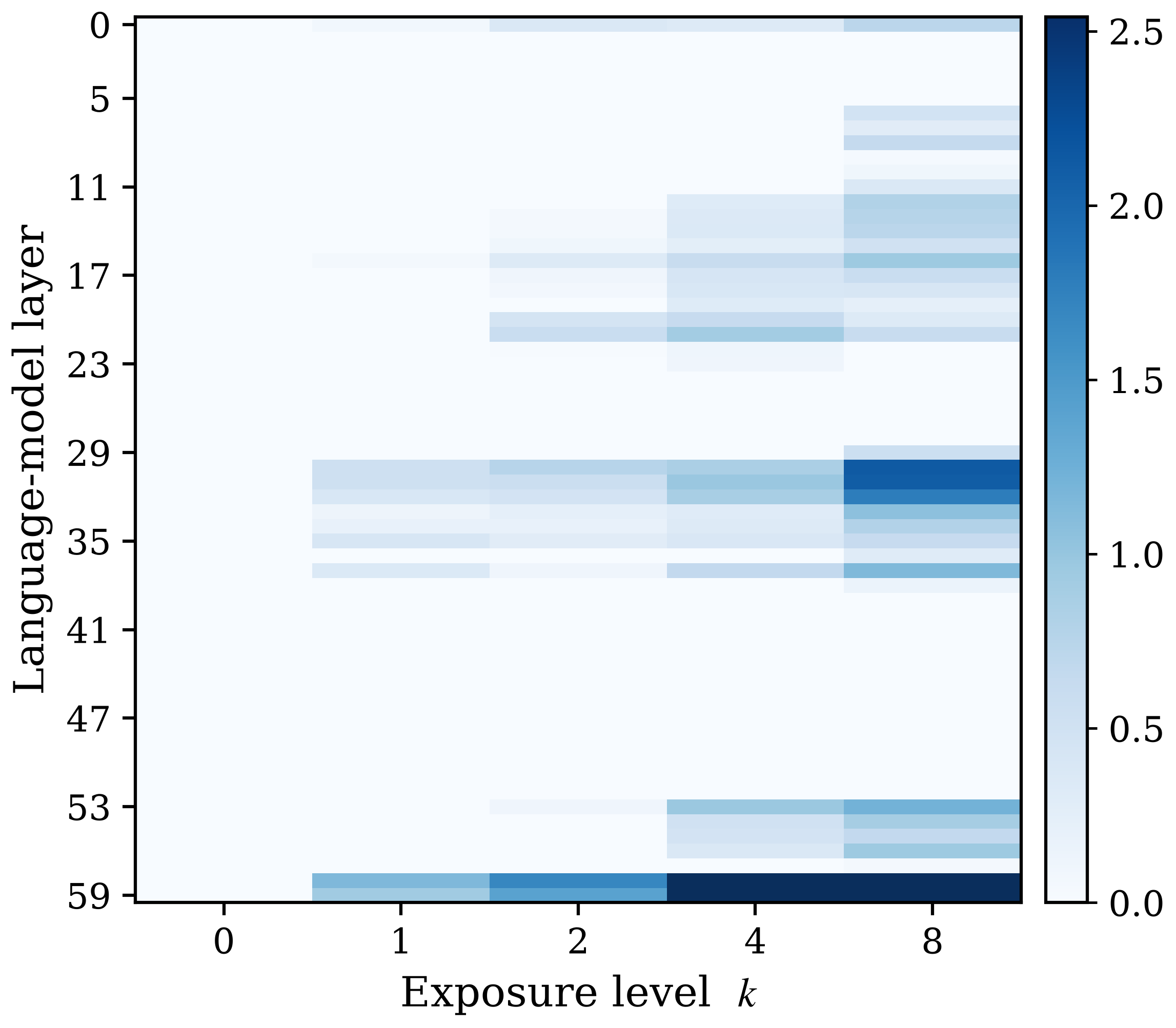}
        \caption{Gemma-4-31B-it, Text.}
        \label{fig:layerwise_31b_text}
    \end{subfigure}
    \hfill
    \begin{subfigure}[t]{0.235\textwidth}
        \centering
        \includegraphics[width=\linewidth]{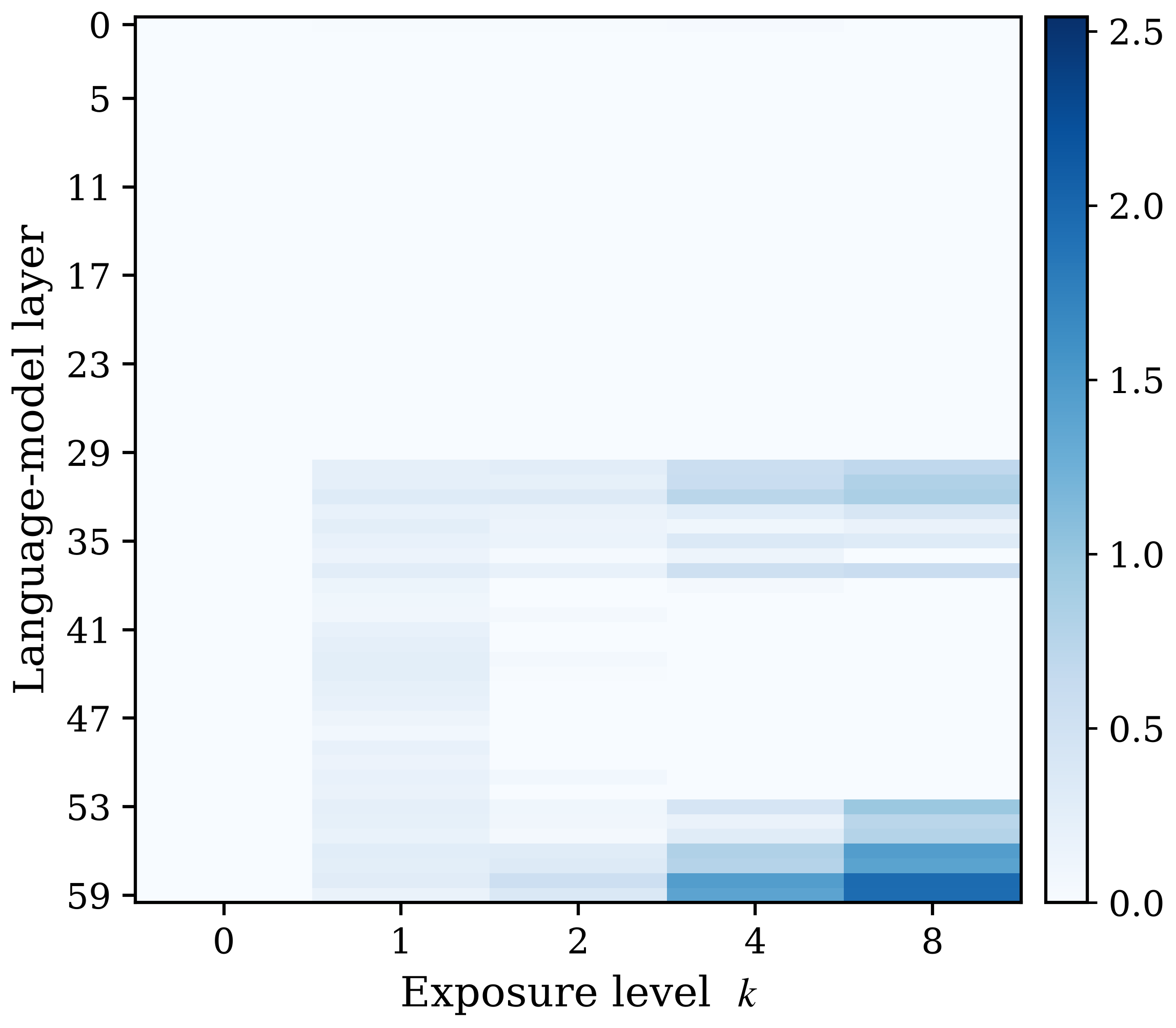}
        \caption{Gemma-4-31B-it, Visual.}
        \label{fig:layerwise_31b_visual}
    \end{subfigure}

    \caption{
        Additional Layer-Wise Readouts of Exposure-Dependent Action Evidence.
        Each heatmap reports the mean change in distractor action evidence relative to the corresponding $k=0$ condition across model depth.
    }
    \label{fig:layerwise_additional_models}
\end{figure*}

\paragraph{Exposure-Dependent Readout.}
The mean change in distractor evidence relative to zero exposure is
\begin{equation}
H_r(l,k)
=
\frac{1}{n}
\sum_{i=1}^{n}
\left[
S_{i,l}(a_i^{D};k,r)
-
S_{i,l}(a_i^{D};0,r)
\right],
\label{eq:layerwise_heatmap}
\end{equation}
where $n=165$. The readouts show stronger exposure-dependent distractor evidence under Text in the later layers. Figure~\ref{fig:layerwise_additional_models} reports additional results for Qwen3.5-27B and Gemma-4-31B-it. For Qwen3.5-27B, the measured evidence remains weak through much of the network and increases primarily in the final layers. Weak early-layer readouts do not establish the absence of exposure-related information in those representations.

\subsubsection{Matched Component Intervention}
\label{app:component_patching}

We use matched activation interventions to test how selected component outputs affect the exposure-related change in distractor score.

\paragraph{Intervention Protocol.}

For each state and representation, we cache component activations from a high-exposure donor pass ($k=8$). During the matched zero-exposure host pass ($k=0$), the target component outputs at the action-sequence positions are replaced with the cached donor activations.

Using the summed action-token log-probability $\widetilde{S}$, the original exposure effect is
\begin{equation}
G_{i,r}^{D}
=
\widetilde{S}_i(a_i^{D};8,r)
-
\widetilde{S}_i(a_i^{D};0,r).
\label{eq:patching_exposure_effect}
\end{equation}
For a patched component $c$, the recovered score change is
\begin{equation}
R_{i,r,c}^{D}
=
\widetilde{S}_i^{\mathrm{patch}(c)}(a_i^{D};0,r)
-
\widetilde{S}_i(a_i^{D};0,r),
\label{eq:patching_recovery}
\end{equation} 
where $\widetilde{S}_i^{\mathrm{patch}(c)}$ denotes the summed action-token log-probability from the patched host pass. Aggregate recovery is expressed as
\begin{equation}
\mathrm{Recovery}_{r,c}
=
100\,
\frac{\mathbb{E}_i[R_{i,r,c}^{D}]}
{\mathbb{E}_i[G_{i,r}^{D}]}.
\label{eq:patching_recovery_ratio}
\end{equation}
This ratio is defined when the mean original effect is nonzero and is not constrained to the interval $[0,100]$. In the state-level scatter plots, the horizontal coordinate is $G_{i,r}^{D}$ and the vertical coordinate is $R_{i,r,c}^{D}$. The diagonal indicates full recovery of the original score change, and the horizontal zero line indicates no recovered score change.

\paragraph{Component Comparisons.}
\begin{wrapfigure}[20]{r}{0.36\columnwidth}
    \centering
    \vspace{0em}
    \includegraphics[width=\linewidth]{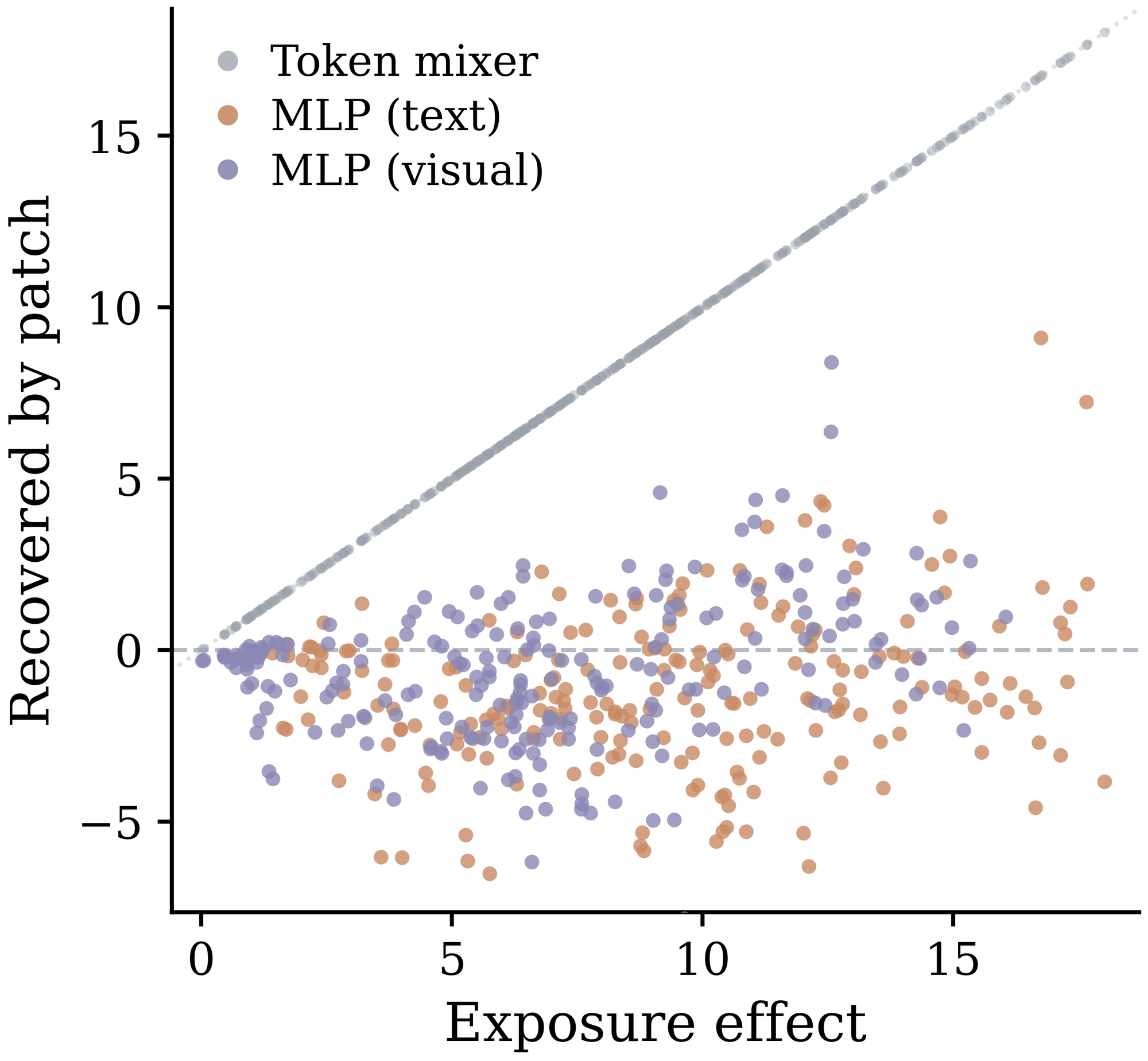}
    \caption{
        Additional Qwen3.5-27B Results for the Matched Component Intervention.
        Each point denotes one ALFWorld decision state.
        The diagonal denotes full recovery, while the horizontal dashed line denotes zero recovered score change.}
    \label{fig:component_patching_27b}
    \vspace{0em}
\end{wrapfigure}
We use the cross-position mixing intervention as a full-recovery reference and the MLP intervention as the primary diagnostic comparison. In Qwen3.5-9B, transplanting high-exposure MLP outputs produces a score change opposite to the original exposure effect under both representations. In Gemma-4-31B-it, the same intervention produces a change in the same direction as the original effect. These findings describe the effects of the specified activation transplants. Additional Qwen3.5-27B results show the same qualitative MLP intervention pattern as Qwen3.5-9B and are presented in Figure~\ref{fig:component_patching_27b}.

\subsection{Examples of Rendered Visual Skills}
\label{app:rendered_visual_skills}

\begin{figure*}[t]
    \centering
    \includegraphics[
        width=\textwidth,
        height=0.82\textheight,
        keepaspectratio
    ]{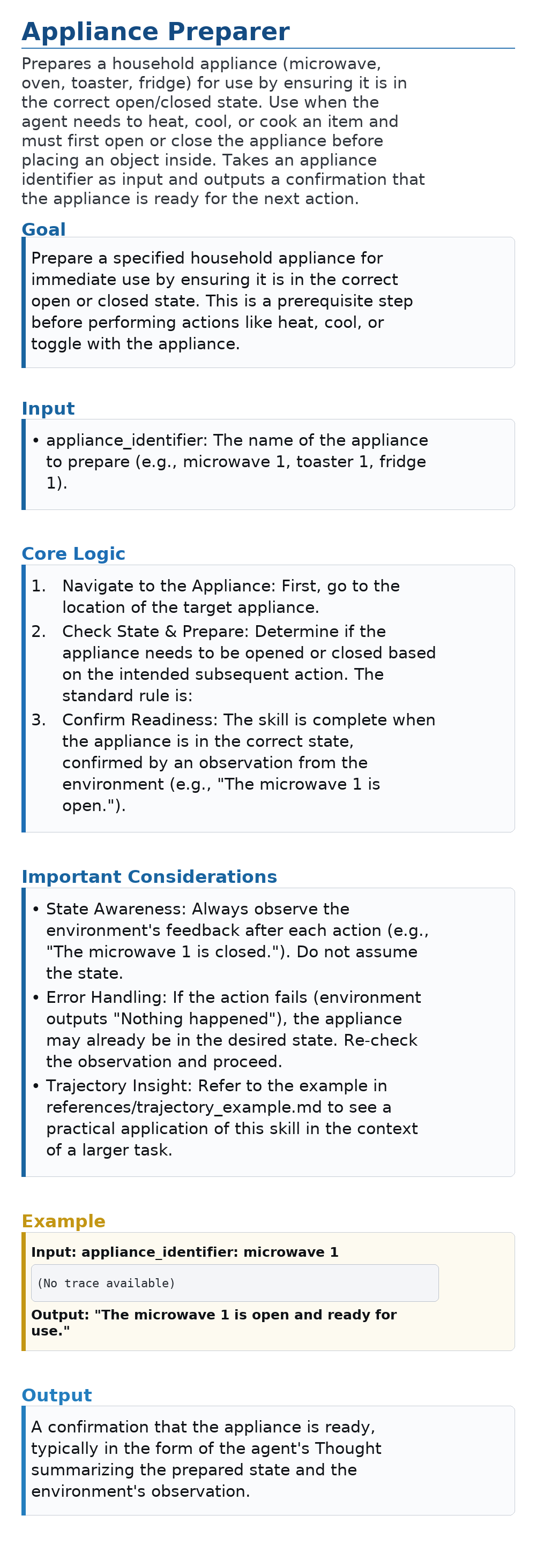}
    \caption{
        Case Study of a Rendered Visual Skill from ALFWorld.
        The \textit{Appliance Preparer} skill visually segregates procedural content into explicit functional regions, mapping operational conditions, execution steps, and expected outputs into distinct spatial blocks.
    }
    \label{fig:rendered_skill_alfworld}
\end{figure*}

\begin{figure*}[t]
    \centering
    \includegraphics[
        width=\textwidth,
        height=0.82\textheight,
        keepaspectratio
    ]{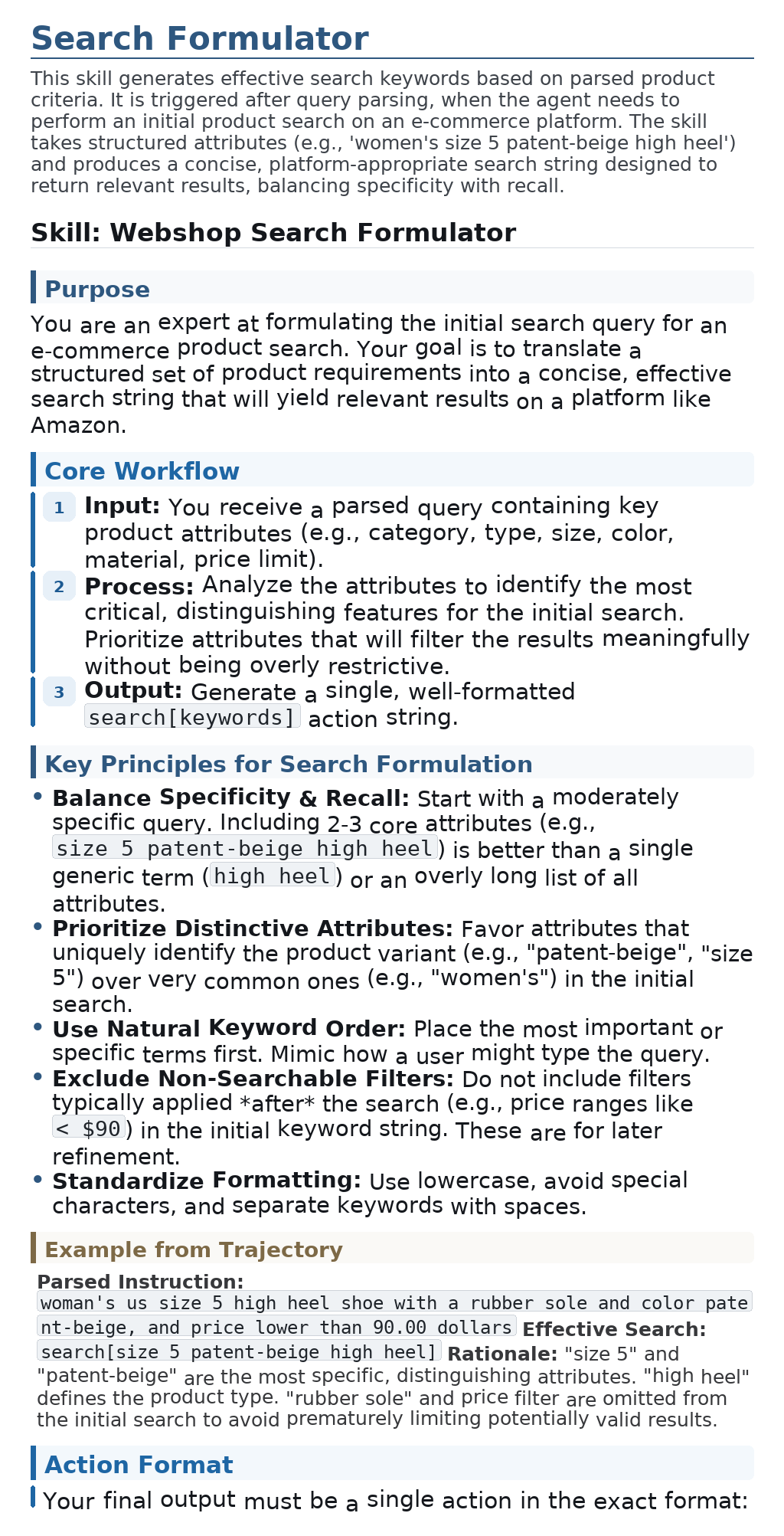}
    \caption{
        Case Study of a Rendered Visual Skill from WebShop.
        The \textit{Search Formulator} skill presents product-parsing procedures and output-format constraints in a structured visual layout.
    }
    \label{fig:rendered_skill_webshop}
\end{figure*}

Figures~\ref{fig:rendered_skill_alfworld} and~\ref{fig:rendered_skill_webshop} illustrate the structured visual format used to present procedural content. The ALFWorld example organizes an appliance-related skill into functional regions, while the WebShop example presents product-search procedures and output constraints.

\end{document}